# Using Large Language Models to Construct Virtual Top Managers: A Method for Organizational Research

Antonio Garzon-Vico, Krithika Sharon Komalapati, Arsalan Shahid, Jan Rosier

*University College Dublin*

## Abstract

This study introduces a methodological framework that uses large language models to create virtual personas of real top managers. Drawing on real CEO communications and Moral Foundations Theory, we construct LLM-based participants that simulate the decision-making of individual leaders. Across three phases, we assess construct validity, reliability, and behavioral fidelity by benchmarking these virtual CEOs against human participants. Our results indicate that theoretically scaffolded personas approximate the moral judgements observed in human samples, suggesting that LLM-based personas can serve as credible and complementary tools for organizational research in contexts where direct access to executives is limited. We conclude by outlining implications for future research using LLM-based personas in organizational settings.

**Keywords:** Large Language Models, Virtual Personas, Content Analysis, Upper Echelons.

## Introduction

In organizational research, there is growing interest in understanding the behaviour of top managers, whose actions shape organizational outcomes. Traditional methods—such as interviews, surveys, and observational studies—are often impractical because access to these individuals is limited. To address this challenge, researchers have increasingly turned to content analysis of publicly available CEO communications (Harrison et al., 2019). Such studies demonstrate that content analysis can yield valuable and valid insights into the psychological characteristics of organizational leaders (Duriau et al., 2007), providing a scalable and objective alternative to direct measurement. Yet, while these methods capture meaningful aspects of communication and cognition, they remain imperfect substitutes for studying the individuals themselves—often missing the depth of real human decision-making.

Recent advances in large language models (LLMs) offer a promising way to extend text-analytic approaches and overcome some of their limitations. Because LLMs can generate responses, make judgements, and act as interactive agents, they provide a means of transforming static textual data into dynamic behavioral data. This potential has already been demonstrated in other fields, where LLMs have been used as substitutes for human participants—for example, to reproduce human-like responses in classic experiments (Dillion et al., 2023; Grossmann et al., 2023; Sreedhar & Chilton, 2024). Researchers have also introduced individual variation by constructing fictional personas with psychological scaffolds such as personality traits or values (Ghaffarzadegan et al., 2024; Horton, 2023; Joshi et al., 2025; Lin, 2025) or by developing detailed synthetic backstories (Moon et al., 2024). Others have personalized models using real individuals' demographic attributes (Aher et al., 2023) or social-media texts (Rahimzadeh et al., 2025).Taken together, these efforts indicate a growing potential for LLM-based participants.

Despite these advances, prior work has rarely combined real, individual-level data with theoretically grounded psychological frameworks, leaving open the question of how closely LLM-based personas can approximate real human behaviour. Building on this gap, we develop virtual personas of top managers grounded in personalized content analysis. Rather than relying on fictional traits or demographic prompts, we construct these personas from CEOs' publicly available communications—such as speeches, interviews, and annual reports—capturing context-specific nuances of individual expression. We pair this textual evidence with scores derived from Moral Foundations Theory to produce virtual replicas aligned with the moral profiles of real leaders. This approach extends content analysis methods, which typically classify broad psychological constructs (Fyffe et al., 2024; Rathje et al., 2024; Stavropoulos et al., 2024), by using content analysis to tailor LLM participants to actual individuals. In doing so, our work brings the two streams together—using LLMs not only as substitutes for human participants but also as a way to unlock richer behavioral insight from authentic CEO communications. Finally, by benchmarking these virtual CEOs against human participants (Trott, 2024), we assess the alignment between text-based and LLM-generated moral profiles and evaluate whether these personas reproduce established patterns of moral judgement in sacrificial dilemmas.

## Literature Review

Trained on vast datasets, LLMs can generate human-like responses and simulate social interactions, making them powerful tools for studying decision making in controlled settings (Dillion et al., 2023; Hutson, 2023). They have been used to model voting behaviour (Argyle et al., 2023), explore the development of social-cognitive skills in childhood (Kovač et al., 2023), replicate legal reasoning (Hamilton, 2023), and reproduce human responses in moral dilemmas such as the Trolley Problem (Dillion et al., 2023). Several studies report striking alignment between LLM outputs and aggregate patterns from psychology experiments (Trott, 2024), with some showing correlations as high as 0.95 with mean human responses (Dillion et al., 2023) and others reproducing 76 percent of main effects across 133 marketing experiments (Yeykelis et al., 2024).

Beyond replicating individual decision making, LLMs have also been used to model collective behaviour. They simulate cooperation and competition in the Prisoner's Dilemma (Akata et al., 2025), reproduce human responses in economic scenarios such as the Ultimatum Game (Grossmann et al., 2023), and generate social networks in which agents exchange information and shift attitudes in response to real-world events, including the release of nuclear wastewater in Japan (Gao et al., 2023) and epidemic outbreaks (Williams et al., 2023).

These early approaches, while promising, relied on average modelling—that is, evaluating how well LLMs reproduce aggregate patterns or the "wisdom of the crowd" reflected in human data. Although this strategy captures central tendencies, it overlooks the diversity of individual perspectives that characterize human behaviour. As a result, LLM-generated responses can default to stereotypical patterns (Cheng et al., 2023), fail to align with the traits they are intended to represent (Sun et al., 2023), and flatten important group differences (Wang et al., 2024), raising concerns about the reliability of such models for behavioral research. To make LLM-based autonomous agents viable substitutes for human participants, they must therefore be able to represent individual variability (Ghaffarzadegan et al., 2024).

***Introducing Individual Variance***

LLM-based personas vary widely in the richness of information used to define them. Some studies rely on demographic cues to introduce individuality, using features such as surnames, gender, or ethnicity to simulate social diversity. For example, Aher et al. (2023) assigned demographic attributes to fictional participants and examined whether these simulated individuals could reproduce behavioral patterns observed in human studies. Other work extends this approach by embedding personas in multi-agent environments where they comment, respond, and interact in ways that sometimes include antisocial behaviour. In studies by Park et al. (Park et al., 2023;

Park et al., 2022), the resulting social dynamics were often indistinguishable from interactions involving real human participants.

Other researchers have sought to represent individuality through more psychologically meaningful information—motivations, values, and attitudes—rather than surface-level demographic cues (Lin, 2025). For example, Ghaffarzadegan et al. (2024) created 20 LLM "workers" with distinct personality profiles who decided whether to adopt a green or blue shirt at work; incorporating personality prompts increased the realism of adoption patterns. Horton (2023) similarly endowed LLM personas with specific values and preferences, finding that agents instructed to prioritize equity consistently selected equitable outcomes. Building on this idea, Joshi et al. (2025) introduced the PB&J (Psychology of Behavior & Judgments) framework, which enriches personas by adding psychological scaffolds—LLM-generated rationales based on theories such as the Big Five, Schwartz's values, and Primal World Beliefs. Across these studies, embedding psychological content enables LLM personas to capture greater individual variation and to generate decisions that more closely resemble human reasoning.

Another strand of research deepens individuality by embedding personas in richer narrative or biographical contexts. Moon et al. (2024) introduce the Anthology framework, which conditions LLM personas on synthetic backstories containing biographical details, values, and lived experiences. Likewise, Piao et al. (2025) propose a model of "silicon participants" built around three components: profiles and status (stable and dynamic traits such as age, education, and economic standing), minds (emotions, opinions, and cognitive processes), and social behaviors (mobility, interaction, and economic activity). These approaches suggest that grounding LLM personas in coherent personal histories and internal states enhances both behavioral stability and their resemblance to real human participants.

### *Using Data from Real Individuals*

Extending this gradient of realism, some studies have grounded LLM personas in data from real individuals rather than relying solely on synthetic inputs. The simplest approach uses demographic information to anchor personas within genuine human distributions. For example, Argyle et al. (2023) drew on profiles from the American National Election Studies, and Hewitt et al. (2024) sampled demographic attributes from seventy nationally representative U.S. survey experiments—both finding strong alignment (up to $r = .85$) between simulated and human responses. Using real demographic data therefore improves fidelity relative to fictional personas, providing an initial step toward human-grounded modelling.

Still, relying solely on demographic data cannot capture the richness of human decision-making. A promising direction is to build LLM-based digital versions of real individuals—models that encode human knowledge and experiences to reproduce behaviour within specific populations (Ong, 2024; Park et al., 2023). This approach leverages the vast amount of textual data people already generate, moving beyond traditional text-analytic methods that score predefined constructs or train supervised NLP models on labelled categories (de Kok, 2025; Speer et al., 2024). Instead of using text to measure individuals, LLMs can use it to represent them—transforming the same rich, individual-level data into functioning personas capable of generating behaviour.

Some studies have begun to test this possibility, offering initial evidence that feeding real textual data into LLMs can produce personas that meaningfully reflect the individuals from whom the data are drawn. Blyler and Seligman (2024), for example, asked twenty-six participants to record fifty streams of consciousness each, along with brief biographical profiles. Using these materials, they generated personalized narratives that twenty-five participants judged as mostly or completely accurate, suggesting that the models captured their distinct voices and perspectives. Park et al. (2024) reported similar results when they provided LLMs with participants' demographic data and complete interview transcripts: the models predicted participants' actual survey responses more accurately and exhibited smaller demographic disparities. Extending this logic at scale, Rahimzadeh et al. (2025) created SYNTHIA, a dataset of 30,000 synthetic personas derived from BlueSky users' posting histories, enriched with demographic and temporal information. Collectively, these studies indicate that when LLMs are grounded in authentic human text, they begin to approximate the individuality and behavioral richness of the people they represent.

The studies reviewed above show that LLMs can approximate the individuality of real people when grounded in authentic textual data, yet their ability to interpret such data in psychologically meaningful ways remains limited. Existing approaches capture surface features—linguistic style, topical focus, and demographic context—but they struggle to infer the deeper motives and dispositions that drive behaviour. Real individuals act not only on what they say but on stable values, experiences, and personality traits that shape decisions across contexts. A natural next step is therefore to pair textual inputs with theoretically informed interpretations of what those texts reveal. Established natural language processing tools already enable trait inference by mapping language to psychological constructs (Fyffe et al., 2024); embedding these inferred profiles into LLM prompts may improve the models' capacity to reproduce individual behavioral patterns with greater fidelity.

Building on this premise, our study examines whether LLMs can serve as realistic proxies for individuals who are otherwise difficult to recruit for traditional experiments. We focus on publicly available communications from 181 CEOs in the pharmaceutical and biotechnology industries and use these texts to construct virtual twins capable of expressing moral reasoning. Specifically, we test whether combining established linguistic tools with theoretical frameworks such as Moral Foundations Theory (Haidt et al., 2009)—and integrating textual, demographic, and trait-based inputs—enhances the interpretive depth of LLM personas. By providing models with these enriched profiles, we evaluate how LLMs respond to various inputs and whether this allows them to reproduce psychologically meaningful patterns of moral judgement. Moral Foundations Theory classifies moral attitudes across five dimensions—care/harm, fairness/cheating, loyalty/betrayal, authority/subversion, and purity/degradation—and thus offers a structured basis for translating textual information into interpretable moral traits for each virtual CEO.

## THE PRESENT RESEARCH

Our study is organized into three sequential phases (see Figure 1), each building on the previous to examine the construction and validity of LLM-generated virtual personas of top managers. In Phase 1, we created four versions of 25 CEO personas by varying the information provided to the LLMs. In Phase 2, we examined whether these input configurations influenced the behaviour of the virtual personas in theoretically consistent ways. In Phase 3, we selected the best-performing version identified in Phase 2 and created virtual personas for the full sample of 181 CEOs, then had these personas complete a study previously administered to human participants (Crone & Laham, 2015), enabling a direct comparison between the behaviour of virtual top managers and real individuals.

——- Insert Figure 1 here ——-

## Phase 1: Persona Construction

### *Data Collection and Preprocessing*

We constructed a sub-sample of 25 CEOs using a purposive, stratified design to ensure variation in reputational standing, including individuals publicly characterized as both high and low performers in trade-press rankings, and then randomly sampled additional CEOs from major pharmaceutical firms to broaden coverage. Eligibility required that each person (a) served as CEO of a biotechnology or pharmaceutical company and (b) had sufficient publicly available text for analysis, defined as the presence of data in all three sources: annual report keynotes, recorded interviews (e.g., YouTube), and Google News articles. To maintain confidentiality and protect privacy, the identities of all CEOs remain anonymous. Texts span the period 2000–2025.

We collected three types of material for each CEO: YouTube video interviews, news articles quoting or directly referencing the CEO, and annual reports containing CEO keynote addresses. A detailed summary of the collected data—including counts of videos, news items, and annual reports for each CEO—is provided in Table A1 (Appendix). Demographic and biographical details (e.g., age, gender, education, career background) were gathered from public sources to supplement the textual material. Videos, articles, and reports were retrieved programmatically using specialized tools (e.g., YouTube Data API, newspaper3k, pdfminer.six). After extraction, all texts were cleaned and pre-processed to remove structural noise and standardize linguistic features, ensuring comparability across sources. The resulting datasets were stored in separate data frames for subsequent analysis.

CEO keynotes from annual reports were extracted using pdfminer.six, and news articles were collected via newspaper3k and BeautifulSoup from Google News. YouTubeTranscriptApi was used to obtain transcripts of CEO interviews from YouTube. Pre-processing and data management were conducted in Python, using pandas and NLTK for text cleaning, tokenization, and standardization.

### *Psychological Scaffolding*

We then developed moral–psychological profiles for each CEO using Moral Foundations Theory (Haidt, Graham, & Joseph, 2009), which distinguishes five core moral dimensions: harm, fairness, loyalty, authority, and purity. The texts collected in Step 1 (interviews, news articles, and annual reports) were analyzed with the Moral Strength API to produce foundation-specific scores ranging from 0 to 10. Scores were computed separately for each source and for each document (e.g., each video, article, or keynote) to capture potential contextual variation. Document-level outputs were summarized for each CEO using descriptive statistics (means, medians, minimums, maximums, quartiles, and standard deviations). This produced a profile assigning each CEO a value on each foundation. Full results are provided in Table A2 in the appendix.

### *Persona Generation*

Using the OpenAI Assistants API, we created four versions of each CEO persona by varying the data included in their instructions. The model configuration was set with temperature = 1.0 and top p = 1.0. The first version relied only on demographic information (age, gender, education, professional experience), serving as a baseline given its common use in prior research. These details were embedded directly in the system prompt so that responses reflected a simplified demographic profile (see Figure A1 in the appendix). The second version used only textual data—keynotes, interviews, and news articles—attached through the OpenAI file-search tool, which allowed the assistant to ground responses in the CEOs' own words (see Figure A2 in the appendix). The tool scanned the complete text files and retrieved contextually relevant passages to inform answers.

The third version used only the Moral Foundations Theory scores derived in Step 2, together with demographic information, to construct personas guided exclusively by quantified moral characteristics (harm, fairness, loyalty, authority, purity). These scores were encoded as structured instructions in the system prompt, directing the model's reasoning around each moral dimension (see Figure A3 in the appendix for LLM instruction). The fourth version combined both the MFT profile and the textual corpus, integrating a theoretical scaffold with the original communications. In practice, this involved embedding the MFT profile within the prompt while also linking the raw texts through the file search tool, providing both moral framing and contextual grounding (see Figure A4 in the appendix for LLM instruction).

**Phase 2: Behavioral Validation**

In this phase, we examined whether providing MFT scores as inputs led the virtual CEOs to behave in ways consistent with those moral profiles, and how these patterns differed from versions that did not include MFT information.

***Moral Foundation Questionnaire***

All four persona versions (demographics; demographics + MFT; textual data; textual data + MFT) for each of the 25 CEOs completed the standard Moral Foundations Questionnaire (MFQ; see Figure A5 in the appendix). The MFQ consists of two parts. In Part 1, respondents evaluate the relevance of specific considerations when judging whether an action is morally right or wrong. Each of the 16 items corresponds to one of the five moral foundations—harm, fairness, loyalty, authority, and purity—and is rated from 0 (not at all relevant) to 5 (extremely relevant). Each persona provided both a rating and a rationale. In Part 2, personas rated their agreement with a series of moral statements (1 = strongly disagree to 6 = strongly agree), again covering the five foundations, and were asked to provide rationales for each response. Responses from both parts were aggregated to produce a score from 0 to 30 for each foundation and each persona version (see Figure A6 in the appendix). To enable comparison with the MFT scores generated via the Moral Strength library in Phase 1, MFQ totals were normalized to a 0–10 scale.

***Compare Versions***

We conducted correlation analyses to assess the alignment between the MFQ-derived scores and the MFT scores embedded in each persona. Specifically, we examined the extent to which the moral profiles supplied during persona construction were reflected in the responses generated during the MFQ. Table 1 reports mean differences, variances, Pearson correlations, and bias estimates (limits of agreement) between the MFQ-based scores and the MFT values obtained through content analysis.

——- Insert Table 1 here ——-

Across versions, results indicate partial but meaningful alignment between text-derived profiles and persona behaviour. The MFT-only version consistently showed the strongest correspondence with text-derived scores, particularly for the Harm/Care ($r = .60$, $p = .002$), Fairness/Reciprocity ($r = .64$, $p < .001$), Loyalty/In-Group ($r = .63$, $p < .001$), and Purity/Sanctity ($r = .67$, $p < .001$) foundations. Mean differences in these domains were modest (ranging from 0.79 to 1.99), and variance estimates suggest relatively stable performance across CEOs. However, correlations for Authority/Respect were weaker ($r = .18$, n.s.), indicating less reliable transfer of this foundation into persona behaviour.

By contrast, the Text-only version and Demographic-only version showed little evidence of consistent alignment. Correlations in these versions were generally low or non-significant, and bias estimates indicated wider divergence from the text-derived profiles. The combined MFT + Text version displayed moderate performance, with small but positive correlations for Fairness/Reciprocity ($r = .35$, $p = .083$), Loyalty/In-Group ($r = .38$, $p = .060$), and Purity/Sanctity ($r = .58$, $p = .002$). Nonetheless, its overall alignment was less stable than the MFT-only version. These trends are also visualized in Figure 2, which shows the correlation coefficients by foundation, highlighting the better performance of the MFT-only version.

—-- Insert Figure 2 here —--

Taken together, these results suggest that providing personas with only MFT scores produces the most consistent behavioral alignment with profiles derived from CEO communications, whereas versions relying solely on text or demographic data do not reproduce the same structure. Importantly, while correlations for several foundations were encouraging, absolute agreement (as reflected in mean bias and variance) was imperfect.

**Phase 3: Comparison With Human Participants**

A crucial step in assessing the feasibility of using LLM-based virtual personas in social science research is to benchmark their performance against real human data (Lyman et al., 2025). The ultimate value of such personas lies not only in internal consistency (i.e., behaving in line with their inputs) but also in their ability to approximate the reasoning and behavioral patterns observed in actual human participants.

Because our study relied on Moral Foundations Theory inputs, we selected Crone and Laham's (2015) experiment as a reference point. Their study, conducted with 307 participants, examined the relationship between moral foundations and responses to sacrificial dilemmas, where individuals judge the acceptability of causing fatal harm to one person in order to save multiple others. This design provides a well-established behavioral test of moral trade-offs, allowing us to evaluate whether our virtual CEOs reproduce the same patterns of moral reasoning found in human samples.

### *Create Full Virtual Sample Using Best Performing Version*

For this part of the study, we created 181 virtual CEOs by applying the same procedure described in Phase 1, but this time focusing exclusively on the best-performing model, Version 3 (MFT scores only). To assemble this broader sample, we drew from the Pharmaceutical Companies Directory, identifying 116 companies and compiling the names of CEOs who had served from the year 2000 onwards. The year 2000 was set as a cut-off, as it marks the point at which CEO communications became more consistently available online in the form of annual reports, news articles, and video interviews. From this list, we selected only those CEOs who had served long enough to generate sufficient textual data for analysis. For each of the 181 CEOs, we collected biographical information—age, gender, education, and a professional summary—mirroring some of the variables available in Crone and Laham's (2015) study. We then gathered all publicly available communications for each CEO and computed their MFT scores using the same text-analysis process outlined in Phase 1.

### *Experimental Replication*

Following Crone and Laham's (2015) study the virtual CEOs were presented with six widely used sacrificial dilemmas (Moore et al., 2008) (see Appendix Figure A7). These dilemmas, which ask whether it is morally acceptable to sacrifice one individual to save many others, are a standard paradigm in experimental ethics and have been extensively employed to study moral decision-making. To examine how virtual CEOs handle such tasks, we implemented two different prompt conditions. In the first, which we call trait-isolated prompting, CEOs were instructed to rely exclusively on their score for a single trait when responding to dilemmas related to that trait (e.g., when judging a harm-related question, they were told to answer solely on the basis of their harm score). In the second, which we call integrated-trait prompting, no such restriction was imposed: the CEOs were asked to respond to all dilemmas by drawing on the full set of trait scores provided. This design allowed us to test two possible ways in which human participants might use their moral dispositions when responding to such questionnaires—either by focusing narrowly on one dimension at a time or by integrating across multiple moral foundations.

For both prompt conditions, we computed an overall sacrifice rating by averaging responses across the six dilemmas (See Appendix Figure A8 for instructions). In addition, the virtual CEOs completed the 30-item Moral Foundations Questionnaire, providing individual-level foundation scores (as in Phase 2, but this time for all 181 virtual CEOs). Following Crone and Laham's (2015) we first examined bivariate correlations between the five foundations and sacrifice ratings and then conducted multiple regression analyses to assess the relative contribution of each foundation to sacrificial judgments.

***Check for stochasticity***

Because LLMs generate responses probabilistically, identical prompts can yield different outputs across runs. To account for this stochasticity, we repeated each of the two prompt conditions five times for all virtual CEOs (see Tables A3.1–A3.5 and A6.1–A6.5 in the appendix). This procedure allowed us to evaluate reliability across replications rather than relying on a single realization of the model's output. We report where variability was observed and present averaged results in Tables 2.1 and 3.1.

To further assess stability, we estimated OLS and Ridge regressions for each prompt condition across all runs, examining coefficient stability and model fit ($R^2$). These analyses tested whether the LLM-generated associations between moral foundations and sacrificial choices were structurally consistent across stochastic runs, robust to modelling choices, and predictive of sacrifice variance (see Tables 2.2, 2.3, 3.2, and 3.3).

---- Insert Tables 2.1,2.2,2.3 and Tables 3.1,3.2,3.3 here ----

To benchmark our virtual CEO sample against human participants, for both prompt conditions we first, reproduced the descriptive tables reported by Crone and Laham (2015), presenting side-by-side results for the human sample and our virtual CEOs (trait isolated and integrated trait). This allowed direct comparability of bivariate correlations and regression estimates. Second, we formally tested for differences between the two samples: for correlations, we used Fisher's r-to-z transformations (Table 2.2,3.2 ) to assess whether the strength of associations between moral foundations and sacrificial judgments differed significantly, and for regression models, we employed Chow tests (Table 2.3,3.3) to evaluate whether the predictive weight of the five moral foundations varied across samples. Third, we complemented these statistical analyses with qualitative interpretation, focusing on whether the theoretical patterns observed in humans—such as Harm and Purity negatively predicting sacrificial acceptability, and Ingroup positively predicting it—were replicated by the virtual CEOs. Together, these steps provide a rigorous assessment of the extent to which LLM-based personas approximate the psychological patterns found in human participants.

***Compare with Human Participants***

**Trait-Isolated Prompting.** In the first set of analyses, we examined the performance of virtual CEOs under the trait-isolated prompting condition, where responses to each dilemma were guided solely by the corresponding foundation score. Across the five runs (see Tables A3.1 to A3.5 in appendix), the descriptive statistics for the five foundations and sacrifice ratings were highly consistent, suggesting that the overall scale of responses was relatively stable across trials. The bivariate correlations indicated that Harm and Fairness were sometimes linked to sacrificial judgments, while Loyalty, Authority, and Purity showed weaker or inconsistent relationships. However, when examining the multiple regression coefficients, which test the unique predictive contribution of each foundation, we observed considerable variability across runs. In some cases, Harm emerged as a positive predictor (Tables A3.1 in appendix), while in others it was negative or null (Tables A3.2 to A3.5 in appendix). Similarly, Purity occasionally displayed negative associations with sacrificial ratings (e.g., Tables A3.3, A3.5) but was not significant in other runs. This inconsistency suggests that while the models reproduce familiar foundations-to-sacrifice patterns in certain instances, the results are not robust across repetitions.

To probe the stability further, we compared ordinary least squares (OLS) and Ridge regression coefficients across runs (see Tables A4.1 to A4.3 in appendix). Here, Authority stood out as the only predictor with consistent directionality across all runs (positive in both OLS and Ridge), whereas Harm, Fairness, Loyalty, and Purity shifted in sign and magnitude between trials. The coefficient stability summary (Table A4.1) confirms this impression: Authority was positive in 5/5 runs, while other foundations alternated between positive and negative effects. The standard deviations of coefficients were also larger than one would expect from human samples tested under identical conditions, underscoring the role of stochasticity in LLM-based personas. Model fits were modest across runs ($R^2$ between .017 and .051 for OLS, see Table A4.2), consistent with prior findings that moral foundations explain only a small portion of variance in sacrificial dilemmas among humans.

Taken together, these results indicate that trait-isolated prompting yields outputs that broadly approximate human-like relationships between moral foundations and sacrificial judgments but lack stability across repeated runs. The preservation of some theoretical signals (e.g., Authority's positive effect, occasional negative Purity effects) shows promise, yet the inconsistent contributions of other foundations limit confidence in this condition as a reliable benchmark.

When benchmarked against Crone and Laham's (2015) human sample, the trait-isolated prompting condition produced several notable parallels. Across the five runs (Tables A4.1 to A4.3 in Appendix), the virtual CEOs showed stable descriptive statistics and a coherent structure of moral foundations. The inter-correlations among foundations were consistently positive—for example, Fairness correlated with Authority ($r = .21–.36$) and Loyalty ($r = .17–.32$) across runs—mirroring the pattern observed in Crone and Laham's human data, where the moral foundations were also interrelated rather than independent. This indicates that the LLM-based personas maintained an internally consistent moral schema similar to that of human participants.

The relative ordering of foundation effects was also broadly preserved. In Crone and Laham's human sample, Harm and Purity showed the strongest negative associations with sacrificial acceptability ($\beta \approx -.25$ in both cases), while Fairness and Authority exhibited small positive coefficients and Loyalty a modest positive effect ($\beta = .21$). The virtual CEOs reproduced this qualitative pattern in several runs (Tables A4.1 to A4.3 in Appendix)): Purity often retained a negative sign (especially in Runs 3–5), and Authority remained consistently positive across all replications. Although these coefficients were smaller in magnitude, the same moral hierarchy

emerged—Harm and Purity constraining sacrificial acceptance more strongly than Fairness, Authority, or Loyalty promoted it.

Formal statistical comparisons supported this pattern of partial convergence. The Fisher's r-to-z tests (Table 3.2) showed that only the correlation between Harm and sacrificial judgments differed significantly between samples ($z = 2.15$, $p = .031$), while differences for Fairness, Loyalty, Authority, and Purity were nonsignificant. Similarly, the Chow tests (Table 3.3) identified a significant difference in the regression coefficient for Harm ($F(1) = 4.88$, $p = .027$) and a marginal one for Loyalty ($F(1) = 3.45$, $p = .063$), suggesting that the LLMs underestimated the moral weight humans place on avoiding harm.

Taken together, the trait-isolated prompting results (Appendix Tables A3.1-A3.5, Tables 3.2, 3.3) indicate that LLM-based virtual CEOs reproduce several structural properties of human moral reasoning—stable inter-foundation coherence, comparable directionality of effects, and similar explanatory power—while showing attenuated moral salience, particularly regarding harm aversion. The trait-isolated condition thus appears to capture the cognitive framework of human moral judgment more accurately than its emotional intensity.

**Integrated-Trait Prompting.** Under the integrated-trait prompting condition, where virtual CEOs drew simultaneously on all moral foundation scores to evaluate each dilemma, the models displayed more stable and coherent moral patterns than under trait-isolated prompting. Across the five runs (Appendix tables A5.1-A5.5), the descriptive statistics and correlation structures were highly consistent: all five foundations were positively intercorrelated (e.g., Fairness–Authority $r = .28–.72$; Fairness–Loyalty $r = .22–.50$), closely mirroring the interdependence reported in Crone and Laham's (2015) human sample. This internal coherence indicates that integrated prompting produced a more globally balanced moral profile, consistent with how human moral intuitions are activated jointly rather than in isolation.

At the regression level, the relationships between moral foundations and sacrificial judgments were relatively modest but directionally consistent across runs. Harm and Purity generally carried weak or near-zero coefficients (e.g., $\beta$Harm = −.11 to +.04; $\beta$Purity = −.04 to +.04), whereas Loyalty and Fairness tended to show small positive effects ($\beta \approx$ .02–.11). The coefficient stability summary (Appendix Table A6.1) confirmed this pattern: Loyalty was positive in all five runs, Fairness in four out of five, and Authority negative in four out of five. This convergence, combined with low variability in standard deviations ($\beta$sd ≈ .03–.07), suggests a more robust signal structure than in the trait-isolated condition. Model fit indices were low but consistent across replications ($R^2$OLS = .003–.015; Appendix Table A6.3), comparable to those observed in human samples where moral foundations typically explain limited variance in sacrificial judgments.

When benchmarked directly against Crone and Laham's human data, the Integrated-Trait Prompting condition again revealed strong structural alignment but attenuated magnitude. The Fisher's r-to-z tests (Table 3.2) indicated a significant difference for the Harm–sacrifice correlation ($z = 2.22$, $p = .026$) and a marginal difference for Purity ($z = 1.79$, $p = .073$), while all other foundations showed no significant divergence. Similarly, Chow tests (Table 3.3) revealed a marginally weaker Harm coefficient ($F(1) = 3.37$, $p = .067$) and a significant difference for Purity ($F(1) = 4.20$, $p = .041$). These results imply that, although the overall configuration of relationships among moral foundations was preserved, the virtual CEOs displayed reduced aversion to both harm and purity violations, replicating the direction but not the full intensity of human moral sensitivity.

Taken together, the Integrated-Trait Prompting results (Appendix Tables A5.1-A5.3, 3.2, 3.3) demonstrate that integrated prompting improves internal consistency and cross-run stability, capturing a richer and more psychologically plausible moral reasoning pattern than trait-isolated prompting. However, as with the trait-isolated prompting condition, the LLM-based personas understate the emotional salience of sacrificial trade-offs—possibly reflecting either limitations in affective simulation or the more utilitarian orientation of the CEO population being modelled.

## Discussion

This study examines whether LLMs can function as realistic replacements for human participants in behavioral research. By constructing personas of top managers grounded in publicly available CEO communications and moral–psychological profiles, we assess the feasibility and validity of using LLM-based participants in organizational studies. Across three phases, we develop, test, and benchmark these virtual CEOs against human participants to evaluate whether such models reproduce the structure of moral reasoning observed in empirical work. We do not position LLMs as substitutes for human respondents but as complementary tools that enable behavioral research with populations that are otherwise difficult to access. Our work provides a methodological template that integrates theoretical grounding, empirical benchmarking, and stochastic testing, allowing researchers to evaluate construct validity and reliability in line with best-practice recommendations emphasizing transparency, replicability, and rigorous validation as foundations of methodological credibility (Aguinis et al., 2021).

## Contribution

Our work has direct implications for organizational research, particularly in the study of top managers where empirical work is constrained by limited access, small and selective samples, and strong self-presentation concerns (Aguinis et al., 2013; Cycyota & Harrison, 2006). By grounding virtual CEO personas in records of leaders' own communications and using a well-specified moral framework to interpret those materials for persona construction, our approach offers a complementary "silicon sample" that maintains a traceable link to real actors while being more tractable for large-scale and experimental research. In this sense, LLM-based personas provide a scalable way of examining theoretically informed representations of executive decision making alongside other existing methods.

Despite our model not being intended as a replacement for real participants, the type of LLM personas we propose here offers several advantages over human respondents: they do not experience fatigue, inattentiveness, or social-desirability pressures, and they can provide explicit rationales for their responses. As a consequence, our approach expands the range of empirical questions that organizational researchers can address. Because virtual top managers can be repeatedly queried, scholars can examine phenomena—such as ethical trade-offs—that are often difficult to study with real executives.

In creating our LLM personas from CEOs' own communications, we extend the role of text-analytic methods in organizational research. Traditional linguistic approaches can infer traits, values, or cognitive tendencies from executives' public messages (e.g., Harrison, 2019), but these techniques are essentially static: they extract information without enabling researchers to examine how inferred psychology unfolds in concrete decisions. By pairing textual evidence with generative models capable of responding to realistic dilemmas, our approach turns static profiles into dynamic behavioral simulations. This makes it possible not only to describe leaders' expressed orientations but also to examine how those orientations might shape behaviour in controlled scenarios—an analytical step that conventional content analysis cannot support (Harrison et al., 2019; Stavropoulos, Crone, & Grossmann, 2024; Rathje et al., 2024).

By combining text-analytic approaches with recent developments in LLM-based research in the social sciences, our study offers a significant advance over existing work (Argyle et al., 2023; Yeykelis et al., 2024), particularly those drawing on textual data from specific individuals (Blyler & Seligman, 2024; Park et al., 2024; Rahimzadeh et al., 2025). We respond to recent calls for methods that capture not only aggregate patterns but also individual variability (Ghaffarzadegan et al., 2024; Lin, 2025).

A central methodological contribution is showing that theoretical scaffolding—in our case Moral Foundations Theory—improves behavioral fidelity beyond what can be achieved with text or demographic data alone. This supports the emerging view that interpretive structure, alongside data quantity, is essential for realism (Joshi et al., 2025). Embedding moral dimensions derived from real CEO communications constrains the model's reasoning in line with psychological theory, thereby strengthening construct validity. Our comparison of model configurations in phase 2 shows that the MFT-only version yields the most stable and interpretable alignment between input traits and behavioral outputs, underscoring the value of theoretically grounded design and revealing the limits of using textual data on their own, which in our case introduced behavioral noise.

Another key contribution lies in benchmarking our virtual CEOs against human participants (Phase 3), providing empirical validation largely absent in earlier LLM-based studies. The observed pattern—Harm and Purity negatively predicting sacrificial acceptability, Authority positively—indicates that our personas reproduce the structure of human moral reasoning. At the same time, their attenuated affective intensity produced a distinct "CEO flavor." We view this not as a limitation but as a realistic reflection of elite managerial cognition. Prior work in strategic and behavioral ethics suggests that executives often rely on calculative, outcome-focused reasoning and express moral concerns in restrained, impersonal terms (Haidt & Kesebir, 2010; Martin et al., 2021; Reynolds, 2006; Tenbrunsel & Smith-Crowe, 2008). This underscores the value of demographic cues for enhancing realism: such cues may help the model capture this professionalized reasoning style and differentiate it from the general-population sample used in other prior studies (Crone and Laham, 2015). Methodologically, this shows how demographic grounding can work alongside theoretical scaffolding to produce psychologically credible simulations of specific populations.

Building on this benchmarking exercise, our results also clarify how different prompting strategies shape the behavioral reliability of LLM-based personas. When moral foundations were activated in isolation, virtual CEOs reproduced several structural features of human moral reasoning but showed substantial instability across stochastic runs, with only Authority exhibiting consistent effects. This pattern suggests that isolating traits can recover theoretically meaningful signals but does so unreliably, making inference sensitive to model variance. In contrast, integrated-trait prompting—where all moral foundations jointly informed each judgement—produced markedly greater stability and internal coherence. Under this condition, moral foundations were consistently intercorrelated, regression coefficients showed less fluctuation across runs, and the overall configuration of effects more closely mirrored human data, albeit with attenuated magnitude. Methodologically, this comparison indicates that LLM personas behave more like human decision-makers when theoretical traits are embedded as an integrated profile rather than activated independently.

## Limitations and Future Directions

Our study also highlights several limitations that point to promising avenues for future research. Although stochastic variation is often treated as noise, our findings indicate that part of this variation may reflect the natural inconsistency of human judgement (Rehren & Sinnott-Armstrong, 2022). From this perspective, variation across runs can mirror the intra-individual fluctuations documented in repeated human measures. However, while averaging across runs reduces random noise, greater standardization is needed to ensure reproducibility. Researchers could therefore employ ensemble prompting (issuing multiple, semantically equivalent prompts and aggregating responses), maintain parameter repositories that record temperature, top-p values and random seeds, and use model-level variance controls when available. A particularly promising direction is the use of ensemble agents, where several instances of the same persona deliberate before a final answer is produced, providing more stable and theoretically grounded results.

One of our findings shows that models using MFT alone performed better than those incorporating textual data. The inclusion of text appears to introduce noise, suggesting that LLMs may not yet generate theoretically meaningful behavioral insights without structured interpretive guidance. In other words, current models lack the analytic tools that human researchers apply when using dictionaries or theoretical constructs. Developing interfaces that enable models to analyze their own textual corpora through predefined theoretical lenses could reduce reliance on researcher framing and lower the labor intensity of the method. For instance, a model could first extract moral statements from its own dataset and then interpret them using an internalized moral framework such as MFT or the Big Five, functioning as a "self-analyzing participant." Coupled with meta-prompting—asking the model to explain its reasoning before responding—this approach could improve transparency, interpretive depth, and theoretical alignment.

A further limitation concerns the relationship between the textual inputs used to construct our personas and the opaque pre-training data of contemporary LLMs. The model we employ has almost certainly been exposed, during pre-training, to some portion of the online material that we later curate as CEO corpora. This creates a potential circularity: the persona may be partly shaped by latent representations of the same leaders already encoded in the model parameters, rather than solely by the structured profiles and documents supplied at inference time. Our design reduces this risk by imposing explicit moral scaffolding and by relying on behavioral tasks and instruments that were not present in the prompting context, yet we cannot guarantee strict independence between training and evaluation data. Future research could address this constraint by using open-weight models trained on known corpora, as well as by masking or counterbalancing information about specific leaders when constructing and testing virtual personas.

The methodological framework introduced here also opens opportunities for experimental applications using multiple agents. Virtual leaders could be placed in simulations of crises, ethical dilemmas, or negotiations to examine decision making under uncertainty without the ethical or logistical constraints of field studies. Extending these simulations to multi-agent settings would enable the study of inter-organizational cooperation and competition, contributing to theories of strategic interaction, moral contagion, and collective intelligence. A natural extension is to examine whether interactions among virtual leaders produce emergent patterns that resemble organizational or industry-level dynamics over time, raising the broader question of how multi-agent environments can illuminate phenomena that are difficult to observe directly in human populations.

A final direction for future research concerns the theoretical scaffolds used to construct virtual personas. Although we relied on Moral Foundations Theory, other psychological frameworks may yield richer or complementary representations of individual decision tendencies. Subsequent studies could examine the effects of alternative scaffolds such as the Big Five, Schwartz's values, or utilitarian–deontological distinctions, or combine multiple frameworks to increase multidimensional realism. Exploring how different theoretical lenses shape behavioural outputs would also clarify whether LLM personas are sensitive to deeper motivational structures or primarily reproduce surface-level cues.

## Conclusion

This study shows that LLMs, when equipped with clear theoretical scaffolds, can serve as credible instruments for simulating individual behaviour in organisational research. By grounding virtual personas in authentic CEO communications and structured moral profiles, we created virtual leaders capable of expressing patterns of moral reasoning that resemble those observed in human participants. Our results suggest that theoretically guided LLM personas offer a complementary approach for studying populations that are difficult to access directly, while also extending text-analytic methods from static measurement toward dynamic behavioural simulation. As human and artificial reasoning continue to converge, such virtual participants provide not only a new methodological tool for organizational research but also a basis for reconsidering how cognition can be modelled within the social sciences.

## References

Aguinis, H., Gottfredson, R. K., & Joo, H. (2013). Best-Practice Recommendations for Defining, Identifying, and Handling Outliers. *Organizational research methods*, *16*(2), 270-301. https://doi.org/10.1177/1094428112470848

Aguinis, H., Hill, N. S., & Bailey, J. R. (2021). Best Practices in Data Collection and Preparation: Recommendations for Reviewers, Editors, and Authors. *Organizational research methods*, *24*(4), 678-693. https://doi.org/10.1177/1094428119836485

Aher, G., Arriaga, R. I., & Kalai, A. T. (2023). *Using Large Language Models to Simulate Multiple Humans and Replicate Human Subject Studies*. https://go.exlibris.link/Rp6XrYH6

Akata, E., Schulz, L., Coda-Forno, J., Oh, S. J., Bethge, M., & Schulz, E. (2025). Playing repeated games with large language models. *Nature Human Behaviour*, 1-11.

Argyle, L. P., Busby, E. C., Fulda, N., Gubler, J. R., Rytting, C., & Wingate, D. (2023). Out of One, Many: Using Language Models to Simulate Human Samples. *Political analysis*, *31*(3), 337-351. https://doi.org/10.1017/pan.2023.2

Blyler, A. P., & Seligman, M. E. (2024). Personal narrative and stream of consciousness: an AI approach. *The journal of positive psychology*, *19*(4), 592-598.

Cheng, M., Durmus, E., & Jurafsky, D. (2023). Marked personas: Using natural language prompts to measure stereotypes in language models. *arXiv preprint arXiv:2305.18189*.

Crone, D. L., & Laham, S. M. (2015). Multiple moral foundations predict responses to sacrificial dilemmas. *Personality and Individual Differences*, *85*, 60-65. https://doi.org/https://doi.org/10.1016/j.paid.2015.04.041

Cycyota, C. S., & Harrison, D. A. (2006). What (Not) to Expect When Surveying Executives:A Meta-Analysis of Top Manager Response Rates and Techniques Over Time. *Organizational research methods*, *9*(2), 133-160. https://doi.org/10.1177/1094428105280770

de Kok, T. (2025). ChatGPT for textual analysis? How to use generative LLMs in accounting research. *Management Science*.

Dillion, D., Tandon, N., Gu, Y., & Gray, K. (2023). Can AI language models replace human participants? *Trends in Cognitive Sciences*, *27*(7), 597-600. https://doi.org/10.1016/j.tics.2023.04.008

Duriau, V. J., Reger, R. K., & Pfarrer, M. D. (2007). A Content Analysis of the Content Analysis Literature in Organization Studies: Research Themes, Data Sources, and Methodological Refinements. *Organizational research methods*, *10*(1), 5-34. https://doi.org/10.1177/1094428106289252

Fyffe, S., Lee, P., & Kaplan, S. (2024). “Transforming” Personality Scale Development: Illustrating the Potential of State-of-the-Art Natural Language Processing. *Organizational research methods*, *27*(2), 265-300. https://doi.org/10.1177/10944281231155771

Gao, Y., Xiong, Y., Gao, X., Jia, K., Pan, J., Bi, Y., Dai, Y., Sun, J., Wang, H., & Wang, H. (2023). Retrieval-augmented generation for large language models: A survey. *arXiv preprint arXiv:2312.10997*, *2*(1).

Ghaffarzadegan, N., Majumdar, A., Williams, R., & Hosseinichimeh, N. (2024). Generative agent-based modeling: an introduction and tutorial. *System dynamics review*, *40*(1), n/a-n/a. https://doi.org/10.1002/sdr.1761

Grossmann, I., Feinberg, M., Parker, D. C., Christakis, N. A., Tetlock, P. E., & Cunningham, W. A. (2023). AI and the transformation of social science research. *Science (American Association for the Advancement of Science)*, *380*(6650), 1108-1109. https://doi.org/10.1126/science.adi1778

Haidt, J., Graham, J., & Joseph, C. (2009). Above and below left–right: Ideological narratives and moral foundations. *Psychological Inquiry*, *20*(2-3), 110-119.

Haidt, J., & Kesebir, S. (2010). Morality. *Handbook of Social Psychology*, *2*, 797-832. https://doi.org/10.1002/9780470561119.socpsy002022

Hamilton, S. (2023). Blind judgement: Agent-based supreme court modelling with gpt. *arXiv preprint arXiv:2301.05327*.

Harrison, J. S., Thurgood, G. R., Boivie, S., & Pfarrer, M. D. (2019). Measuring CEO personality: Developing, validating, and testing a linguistic tool. *Strategic Management Journal*, *40*(8), 1316-1330. https://doi.org/https://doi.org/10.1002/smj.3023

Hewitt, L., Ashokkumar, A., Ghezae, I., & Willer, R. (2024). Predicting results of social science experiments using large language models. *Preprint*.

Horton, J. J. (2023). *Large Language Models as Simulated Economic Agents: What Can We Learn from Homo Silicus?* (Vol. no. w31122). National Bureau of Economic Research. https://go.exlibris.link/Snz48YD8

Hutson, M. (2023). Guinea pigbots. *Science*, *381*(6654), 121-123. https://doi.org/10.1126/science.adj6791

Joshi, B., Ren, X., Swayamdipta, S., Koncel-Kedziorski, R., & Paek, T. (2025). Improving Language Model Personas via Rationalization with Psychological Scaffolds. *arXiv preprint arXiv:2504.17993*.

Kovač, G., Portelas, R., Dominey, P. F., & Oudeyer, P.-Y. (2023). The socialai school: Insights from developmental psychology towards artificial socio-cultural agents. *arXiv preprint arXiv:2307.07871*.

Lin, Z. (2025). Large Language Models as Psychological Simulators: A Methodological Guide. *arXiv preprint arXiv:2506.16702*.

Lyman, A., Hepner, B., Argyle, L. P., Busby, E. C., Gubler, J. R., & Wingate, D. (2025). Balancing Large Language Model Alignment and Algorithmic Fidelity in Social Science Research. *Sociological Methods & Research*, *54*(3), 1110-1155. https://doi.org/10.1177/00491241251342008

Martin, K. D., Kish-Gephart, J. J., & Detert, J. R. (2021). Moral reasoning in the workplace: A review and extension. *Academy of Management Annals*, *15*(2), 287-325. https://doi.org/10.5465/annals.2019.0054

Moon, S., Abdulhai, M., Kang, M., Suh, J., Soedarmadji, W., Behar, E. K., & Chan, D. M. (2024). Virtual personas for language models via an anthology of backstories. *arXiv preprint arXiv:2407.06576*.

Moore, A. B., Clark, B. A., & Kane, M. J. (2008). Who shalt not kill? Individual differences in working memory capacity, executive control, and moral judgment. *Psychol Sci*, *19*(6), 549-557. https://doi.org/10.1111/j.1467-9280.2008.02122.x

Ong, D. C. (2024). GPT-ology, Computational Models, Silicon Sampling: How should we think about LLMs in Cognitive Science? *arXiv preprint arXiv:2406.09464*.

Park, J. S., O'Brien, J., Cai, C. J., Morris, M. R., Liang, P., & Bernstein, M. S. (2023). Generative agents: Interactive simulacra of human behavior. Proceedings of the 36th annual acm symposium on user interface software and technology,

Park, J. S., Popowski, L., Cai, C., Morris, M. R., Liang, P., & Bernstein, M. S. (2022). Social simulacra: Creating populated prototypes for social computing systems. Proceedings of the 35th Annual ACM Symposium on User Interface Software and Technology,

Park, J. S., Zou, C. Q., Shaw, A., Hill, B. M., Cai, C., Morris, M. R., Willer, R., Liang, P., & Bernstein, M. S. (2024). Generative agent simulations of 1,000 people. *arXiv preprint arXiv:2411.10109*.

Piao, J., Yan, Y., Li, N., Zhang, J., & Li, Y. (2025). Exploring Large Language Model Agents for Piloting Social Experiments. *arXiv preprint arXiv:2508.08678*.

Rahimzadeh, V., Monazzah, E. M., Pilehvar, M. T., & Yaghoobzadeh, Y. (2025). SYNTHIA: Synthetic Yet Naturally Tailored Human-Inspired PersonAs. *arXiv preprint arXiv:2507.14922*.

Rathje, S., Mirea, D.-M., Sucholutsky, I., Marjieh, R., Robertson, C. E., & Van Bavel, J. J. (2024). GPT is an effective tool for multilingual psychological text analysis. *Proceedings of the National Academy of Sciences*, *121*(34), e2308950121.

Rehren, P., & Sinnott-Armstrong, W. (2022). How Stable are Moral Judgments? *Rev Philos Psychol*, 1-27. https://doi.org/10.1007/s13164-022-00649-7

Reynolds, S. J. (2006). Moral awareness and ethical predispositions: investigating the role of individual differences in the recognition of moral issues. *J Appl Psychol*, *91*(1), 233-243. https://doi.org/10.1037/0021-9010.91.1.233

Speer, A. B., Perrotta, J., & Kordsmeyer, T. L. (2024). Taking It Easy: Off-the-Shelf Versus Fine-Tuned Supervised Modeling of Performance Appraisal Text. *Organizational research methods*. https://doi.org/10.1177/10944281241271249

Sreedhar, K., & Chilton, L. (2024). Simulating human strategic behavior: Comparing single and multi-agent llms. *arXiv preprint arXiv:2402.08189*.

Stavropoulos, A., Crone, D. L., & Grossmann, I. (2024). Shadows of wisdom: Classifying meta-cognitive and morally grounded narrative content via large language models. *Behavior research methods*. https://doi.org/10.3758/s13428-024-02441-0

Sun, H., Pei, J., Choi, M., & Jurgens, D. (2023). Aligning with whom? large language models have gender and racial biases in subjective nlp tasks. *arXiv preprint arXiv:2311.09730*.

Tenbrunsel, A. E., & Smith-Crowe, K. (2008). 13 Ethical Decision Making: Where We've Been and Where We're Going. *Academy of Management Annals*, *2*(1), 545-607. https://doi.org/10.5465/19416520802211677

Trott, S. (2024). Large Language Models and the Wisdom of Small Crowds. *Open mind*, *8*, 723-738. https://doi.org/10.1162/opmi_a_00144

Wang, A., Morgenstern, J., & Dickerson, J. P. (2024). Large language models that replace human participants can harmfully misportray and flatten identity groups. *arXiv preprint arXiv:2402.01908*.

Williams, R., Hosseinichimeh, N., Majumdar, A., & Ghaffarzadegan, N. (2023). Epidemic modeling with generative agents. *arXiv preprint arXiv:2307.04986*.

Yeykelis, L., Pichai, K., Cummings, J. J., & Reeves, B. (2024). Using Large Language Models to Create AI Personas for Replication, Generalization and Prediction of Media Effects: An Empirical Test of 133 Published Experimental Research Findings. *arXiv preprint arXiv:2408.16073*.

**Figure 1.**

*Three-Phase Framework for Constructing and Validating Virtual Personas Using Large Language Models*

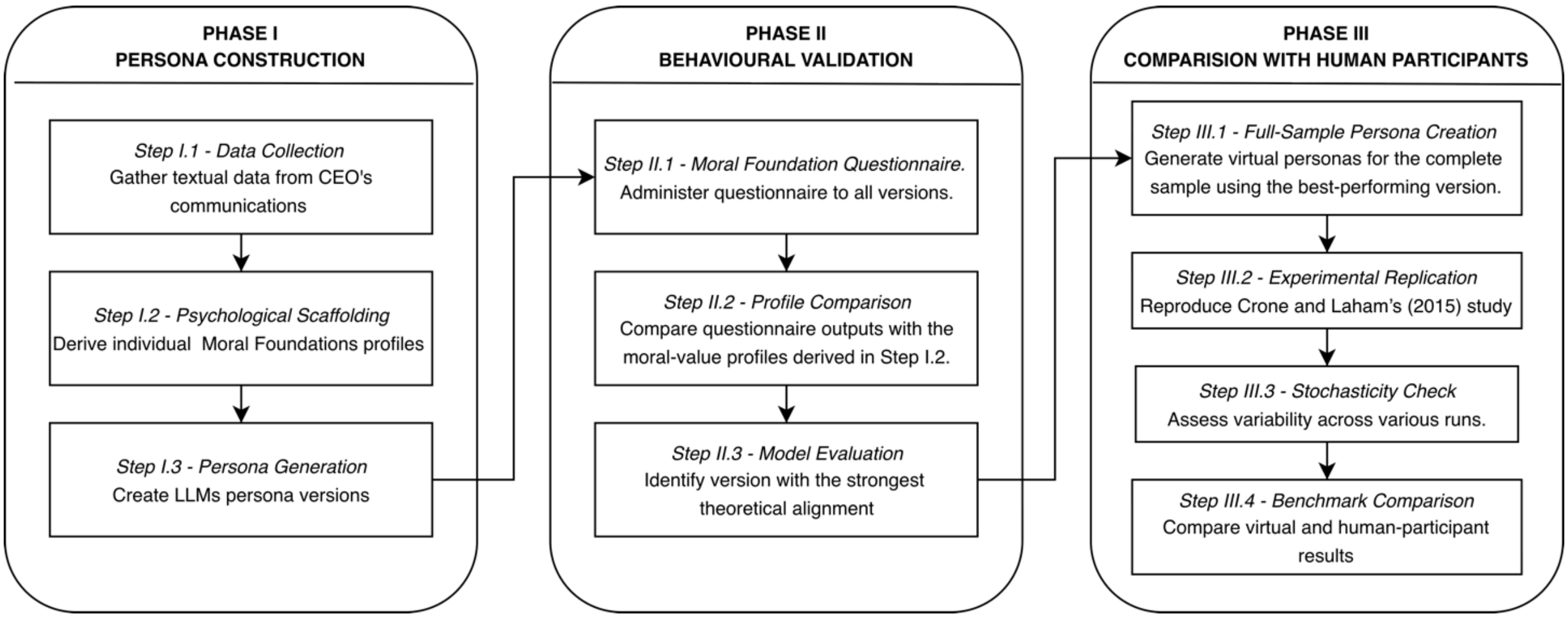

**Table 1**

*Descriptives, correlations, and mean bias (limits of agreement) for four CEO-persona versions, by moral foundation*

| Version | Foundation | MD | Var | Pearson's r (p) | Mean Bias (LoA) |
|---|---|---|---|---|---|
| V1: Demographics | Harm/Care | 1.83 | 0.35 | 0.18 (0.391) | 1.83 (0.65, 3.01) |
| | Fairness/Reciprocity | 1.56 | 0.74 | 0.20 (0.331) | 1.56 (-0.16, 3.28) |
| | Loyalty/In-Group | 0.72 | 0.59 | 0.39 (0.054) | 0.72 (-0.81, 2.25) |
| | Authority/Respect | 0.67 | 0.55 | 0.08 (0.700) | 0.67 (-0.82, 2.15) |
| | Purity/Sanctity | -0.27 | 1.56 | 0.15 (0.464) | -0.27 (-2.76, 2.23) |
| V2: Text | Harm/Care | 2.33 | 1.02 | -0.31 (0.136) | 2.34 (0.31, 4.36) |
| | Fairness/Reciprocity | 2.40 | 0.60 | 0.24 (0.238) | 2.40 (0.85, 3.95) |
| | Loyalty/In-Group | 0.66 | 2.17 | 0.35 (0.085) | 0.66 (-2.29, 3.60) |
| | Authority/Respect | 0.24 | 1.33 | -0.07 (0.732) | 0.24 (-2.07, 2.55) |
| | Purity/Sanctity | -0.28 | 2.39 | -0.10 (0.644) | -0.28 (-3.37, 2.81) |
| V3: MFT | Harm/Care | 1.99 | 0.26 | 0.60 (0.002) | 1.99 (0.97, 3.00) |
| | Fairness/Reciprocity | 1.92 | 0.66 | 0.64 (<0.001) | 1.92 (0.29, 3.55) |
| | Loyalty/In-Group | 1.18 | 0.30 | 0.63 (<0.001) | 1.18 (0.08, 2.27) |
| | Authority/Respect | 1.45 | 0.52 | 0.18 (0.379) | 1.45 (0.01, 2.90) |
| | Purity/Sanctity | 0.79 | 0.61 | 0.67 (<0.001) | 0.79 (-0.78, 2.35) |
| V4: MFT + Text | Harm/Care | 2.16 | 0.59 | 0.04 (0.876) | 2.16 (0.62, 3.70) |
| | Fairness/Reciprocity | 2.00 | 0.63 | 0.35 (0.083) | 2.00 (0.42, 3.58) |
| | Loyalty/In-Group | 1.34 | 0.86 | 0.38 (0.060) | 1.34 (-0.52, 3.20) |
| | Authority/Respect | 1.20 | 0.99 | -0.23 (0.263) | 1.20 (-0.78, 3.19) |
| | Purity/Sanctity | 0.52 | 1.22 | 0.58 (0.002) | 0.52 (-1.69, 2.73) |

Note. Values are based on N = 25 CEOs. Significance is reported via p values in parentheses.

**Figure 2.**

*Pearson correlations with Moral Strength (MSL) by moral foundation across four persona versions*

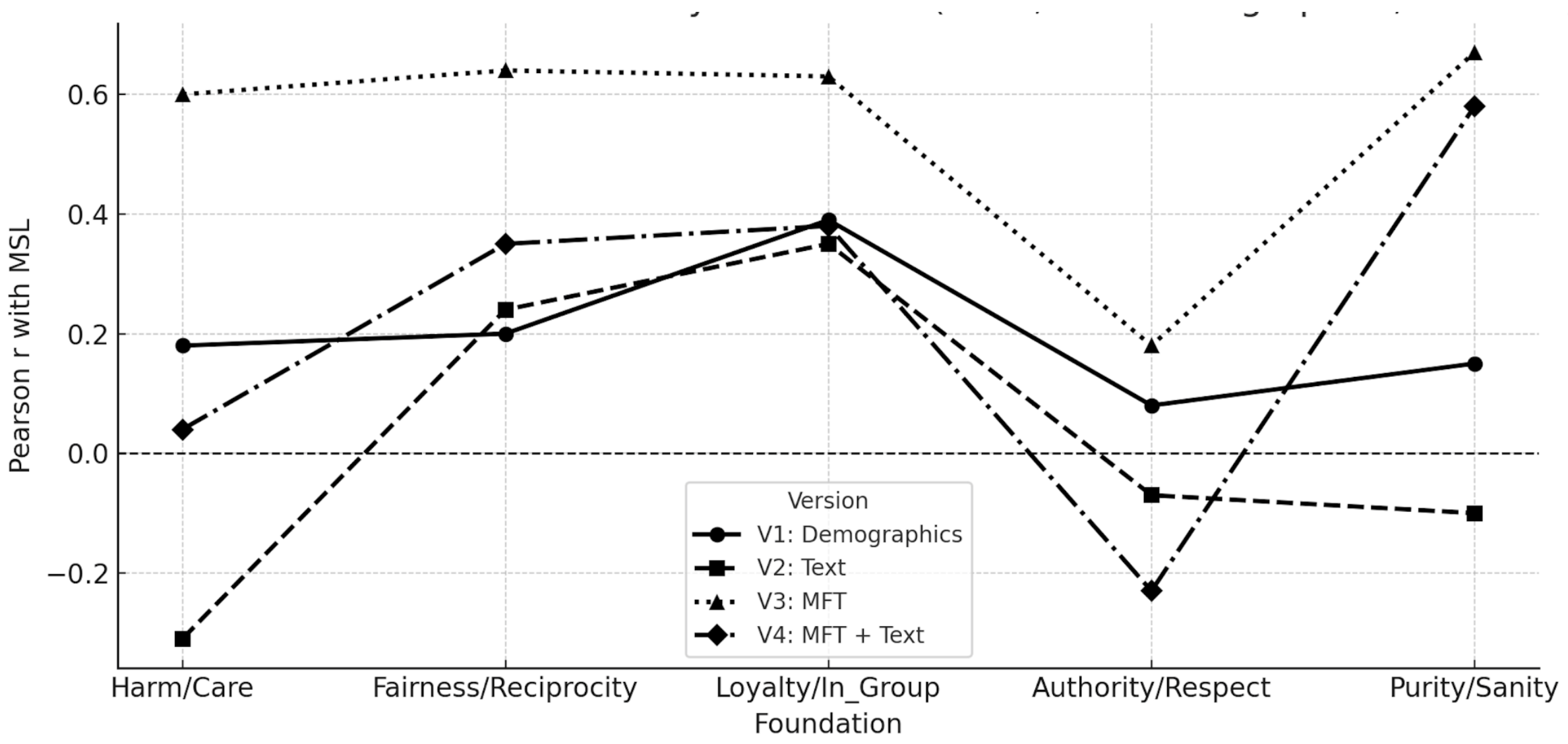

**Table 2.1**

*Mean Results Across Five Runs (Trait-Isolated Prompting)*

| | M | SD | β (SE) | 1 | 2 | 3 | 4 | 5 |
|---|---|---|---|---|---|---|---|---|
| 1. Sacrifice Rating | 2.99 | 0.33 | – | – | | | | |
| 2. Harm | 4.73 | 0.58 | −.00 (.08) | −.01 | – | | | |
| 3. Fairness | 4.55 | 0.48 | −.03 (.08) | −.04 | .19 | – | | |
| 4. Loyalty | 4.45 | 0.42 | −.01 (.08) | 0 | .18 | .26 | – | |
| 5. Authority | 4.45 | 0.34 | .08 (.08) | 0.05 | 0.06 | .15 | .23 | – |
| 6. Purity | 3.50 | 0.72 | −.07 (.08) | −.07 | .14 | .18 | .20 | 0.10 |

Note. Values are means averaged across five stochastic runs. Significance tests are not reported because coefficients are aggregated across replications.

**Table 2.2**

*Fisher's r-to-z Comparison Between LLM Average and Human Benchmark (Trait-Isolated Prompting)*

| Foundation | Pearson r (LLM) | Pearson r (Human) | Fisher's z_diff | Two-tailed p |
|---|---|---|---|---|
| Harm/Care | −0.01 | −0.21 | 2.153 | 0.031 |
| Fairness/Reciprocity | −0.04 | −0.09 | 0.532 | 0.595 |
| Loyalty/In-Group | −0.01 | 0.04 | −0.530 | 0.596 |
| Authority/Respect | 0.05 | −0.04 | 0.954 | 0.340 |
| Purity/Sanctity | −0.07 | −0.16 | 0.967 | 0.334 |

Note. Pearson r values represent correlations between moral foundations and sacrificial judgments for LLM-based personas (averaged across five runs) and the human benchmark sample (Crone & Laham, 2015). Fisher's z_diff tests differences between independent correlations. p values are two-tailed.

**Table 2.3**

*Chow Test of Coefficient Differences Between LLM Average and Human Benchmark (Trait-Isolated Prompting)*

| Foundation | β_Human (SE) | β_LLM (SE) | Δβ (LLM – Human) | $F(1, df_2)$ | p |
|---|---|---|---|---|---|
| Harm/Care | −0.25 (.08) | −0.00 (.08) | 0.25 | 4.883 | 0.027 |
| Fairness/Reciprocity | 0.06 (.08) | −0.03 (.08) | −0.09 | 0.633 | 0.426 |
| Loyalty/In-Group | 0.21 (.08) | 0.00 (.08) | −0.21 | 3.445 | 0.063 |
| Authority/Respect | 0.07 (.09) | 0.08 (.08) | 0.01 | 0.007 | 0.934 |
| Purity/Sanctity | −0.25 (.08) | −0.07 (.08) | 0.18 | 2.531 | 0.112 |

Note. Entries are standardized coefficients (β) with standard errors in parentheses from separate OLS models estimated for the human benchmark sample and the LLM-based personas (averaged across five runs) under the Trait-Isolated prompting condition. Δβ represents the difference between coefficients (LLM − Human). F statistics are from Chow tests of coefficient equality across samples (numerator df = 1). p values are two-tailed. No multiple-comparison correction was applied.

**Table 3.1**

*Mean Results Across Five Runs (Integrated-Trait Prompting)*

| | M | SD | β (SE) | 1 | 2 | 3 | 4 | 5 |
|---|---|---|---|---|---|---|---|---|
| 1. Sacrifice Rating | 2.99 | 0.31 | – | – | | | | |
| 2. Harm | 4.60 | 0.84 | −.03 (.09) | −.01 | – | | | |
| 3. Fairness | 4.29 | 1.08 | .03 (.09) | 0.02 | .33 | – | | |
| 4. Loyalty | 4.46 | 0.56 | .06 (.10) | 0.03 | .38 | .29 | – | |
| 5. Authority | 4.46 | 0.56 | −.04 (.10) | −.01 | .30 | .26 | .62 | – |
| 6. Purity | 3.34 | 0.99 | −.01 (.09) | 0.01 | .33 | .40 | .28 | 0.22 |

Note. Values are means averaged across five stochastic runs. Significance tests are not reported because coefficients are aggregated across replications.

**Table 3.2**

*Fisher's r-to-z Comparison Between LLM Average and Human Benchmark (Integrated-Trait Prompting)*

| Foundation | Pearson r (LLM) | Pearson r (Human) | Fisher's z_diff | Two-tailed p |
|---|---|---|---|---|
| Harm/Care | 0.00 | −0.21 | 2.22 | 0.026 |
| Fairness/Reciprocity | 0.03 | −0.09 | 1.25 | 0.211 |
| Loyalty/In-Group | 0.03 | 0.04 | 0.10 | 0.919 |
| Authority/Respect | 0.01 | −0.04 | 0.52 | 0.603 |
| Purity/Sanctity | 0.01 | −0.16 | 1.79 | 0.073 |

Note. Pearson r values represent correlations between moral foundations and sacrificial judgments for LLM-based personas (averaged across five runs) and the human benchmark sample (Crone & Laham, 2015). Fisher's z_diff tests differences between independent correlations. p values are two-tailed. No multiple-comparison correction was applied.

**Table 3.3**

*Chow Test of Coefficient Differences Between LLM Average and Human Benchmark (Integrated-Trait Prompting)*

| Foundation | β_Human (SE) | β_LLM (SE) | Δβ (LLM – Human) | $F(1, df_2)$ | p |
|---|---|---|---|---|---|
| Harm/Care | −0.25 (0.08) | −0.03 (0.09) | 0.22 | 3.37 | 0.067 |
| Fairness/Reciprocity | 0.06 (0.08) | 0.03 (0.09) | −0.03 | 0.06 | 0.806 |
| Loyalty/In-Group | 0.21 (0.08) | 0.05 (0.10) | −0.16 | 1.17 | 0.279 |
| Authority/Respect | 0.07 (0.09) | −0.04 (0.10) | −0.11 | 0.62 | 0.431 |
| Purity/Sanctity | −0.25 (0.08) | −0.01 (0.09) | 0.24 | 4.20 | 0.041 |

Note. Entries are standardized coefficients (β) with standard errors in parentheses from separate OLS models estimated for the human benchmark sample and the LLM-based personas (averaged across five runs) under the Integrated-Trait prompting condition. Δβ represents the difference between coefficients (LLM − Human). F statistics are from Chow tests of coefficient equality across samples (numerator df = 1). p values are two-tailed. No multiple-comparison correction was applied.

## Appendix

**Table A1**

*Media corpus coverage (counts) for each CEO across three sources*

| CEOs / Media | YouTube Videos | Google News | Annual Reports |
|---|---|---|---|
| CEO 1 | 53 | 126 | 10 |
| CEO 2 | 80 | 154 | 8 |
| CEO 3 | 13 | 77 | 12 |
| CEO 4 | 27 | 32 | 12 |
| CEO 5 | 25 | 45 | 8 |
| CEO 6 | 29 | 30 | 4 |
| CEO 7 | 27 | 13 | 2 |
| CEO 8 | 27 | 103 | 6 |
| CEO 9 | 35 | 5 | 0 |
| CEO 10 | 27 | 56 | 0 |
| CEO 11 | 30 | 77 | 7 |
| CEO 12 | 35 | 6 | 7 |
| CEO 13 | 32 | 23 | 0 |
| CEO 14 | 27 | 7 | 10 |
| CEO 15 | 31 | 13 | 1 |
| CEO 16 | 31 | 3 | 1 |
| CEO 17 | 32 | 50 | 6 |
| CEO 18 | 6 | 5 | 0 |
| CEO 19 | 35 | 71 | 0 |
| CEO 20 | 37 | 27 | 6 |
| CEO 21 | 39 | 83 | 12 |
| CEO 22 | 31 | 48 | 7 |
| CEO 23 | 34 | 32 | 3 |
| CEO 24 | 38 | 57 | 10 |
| CEO 25 | 36 | 39 | 3 |

**Table A2**

*Moral Foundation Theory Score extracted from Moral Strength API for each CEO*

| | Authority (Median, Min, Max) | Harm (Median, Min, Max) | Fairness (Median, Min, Max) | Loyalty (Median, Min, Max) | Purity (Median, Min, Max) |
|---|---|---|---|---|---|
| CEO 1 | 6.86, 4.48, 8.54 | 7.36, 3.18, 8.72 | 6.31, 4.05, 8.19 | 6.63, 5.69, 7.27 | 5.65, 3.48, 8.16 |
| CEO 2 | 6.57, 3.57, 8.27 | 6.21, 2.70, 8.56 | 6.60, 4.44, 7.72 | 6.73, 4.96, 7.92 | 5.73, 2.73, 8.00 |
| CEO 3 | 6.69, 4.7, 8.37 | 6.43, 4.16, 8.48 | 6.71, 4.59, 7.85 | 6.49, 5.59, 7.44 | 4.86, 2.68, 8.13 |
| CEO 4 | 6.43, 5.25, 8.12 | 6.11, 3.03, 8.44 | 6.78, 5.65, 8.05 | 6.68, 5.28, 8.07 | 4.81, 2.80, 7.03 |
| CEO 5 | 6.52, 4.63, 8.25 | 6.48, 2.45, 8.8 | 6.43, 3.92, 7.97 | 6.55, 4.86, 7.83 | 5.19, 2.90, 7.56 |
| CEO 6 | 6.52, 4.01, 7.69 | 6.19, 3.26, 8.65 | 6.71, 5.15, 7.58 | 6.77, 5.24, 8.01 | 5.11, 2.97, 7.96 |
| CEO 7 | 6.45, 4.73, 7.65 | 6.50, 4.56, 7.78 | 6.28, 5.13, 7.43 | 6.47, 5.30, 7.32 | 4.72, 3, 6.5 |
| CEO 8 | 6.73, 5.16, 8.43 | 6.33, 3, 8.66 | 6.76, 5.21, 7.65 | 6.46, 4.79, 7.60 | 5.18, 2.60, 8.11 |
| CEO 9 | 7.12, 5.16, 8.8 | 6.25, 3.93, 8.8 | 8.09, 6.88, 8.51 | 6.45, 5.23, 7.40 | 4.87, 3.47, 5.5 |
| CEO 10 | 6.73, 4.91, 8.20 | 7.02, 2.79, 8.79 | 6.67, 4.05, 8.27 | 6.51, 4.85, 7.55 | 4.74, 2.68, 8 |
| CEO 11 | 6.64, 4.33, 8.55 | 7.15, 3.66, 8.76 | 6.48, 3.75, 8.30 | 6.16, 4.64, 7.70 | 4.89, 2.63, 8.23 |
| CEO 12 | 6.31, 4.98, 7.39 | 6.69, 5.01, 8.27 | 6.55, 5.62, 7.04 | 6.62, 5.47, 7.68 | 5.03, 3.39, 6.31 |
| CEO 13 | 6.74, 4.22, 7.19 | 6.48, 2.41, 8.7 | 7.06, 5.12, 8.58 | 6.29, 4.03, 8.65 | 6.05, 2.66, 8.08 |
| CEO 14 | 6.51, 5.05, 7.44 | 6.73, 4.22, 8.4 | 6.60, 4.8, 8.16 | 6.55, 5.46, 7.33 | 5.24, 2.96, 8.13 |
| CEO 15 | 6.83, 4.48, 8.18 | 5.41, 2.48, 8.29 | 7.04, 4.33, 8.16 | 6.47, 4.51, 7.47 | 5.09, 2.60, 8.08 |
| CEO 16 | 6.84, 6, 7.6 | 6.50, 5.1, 8.7 | 7.73, 6.41, 8.29 | 6.60, 4.84, 7.60 | 7.28, 4.94, 8 |
| CEO 17 | 6.49, 3.52, 8.36 | 6.71, 3.07, 8.8 | 6.69, 4.71, 7.52 | 6.73, 5.66, 7.72 | 4.32, 2.68, 6.40 |
| CEO 18 | 7.15, 5.91, 8.21 | 6.80, 4.92, 8.6 | 7.40, 7.30, 7.49 | 6.56, 5.94, 7.61 | 4.99, 4.20, 5.5 |
| CEO 19 | 6.79, 3.59, 8.73 | 5.50, 2.55, 8.73 | 6.98, 4.08, 8.29 | 6.41, 2.25, 8.21 | 4.97, 2.52, 8.1 |
| CEO 20 | 6.60, 4.98, 8.08 | 6.66, 3.39, 8.56 | 6.29, 3.87, 7.71 | 6.54, 4.32, 7.52 | 4.47, 2.80, 6.98 |
| CEO 21 | 6.40, 3.77, 8.51 | 6.30, 2.58, 8.76 | 6.72, 4.29, 8.71 | 6.47, 4.17, 8.13 | 5.06, 2.78, 8.6 |
| CEO 22 | 6.99, 4.80, 8.32 | 6.83, 3.57, 8.36 | 6.08, 3.92, 7.40 | 6.43, 4.95, 7.56 | 5.20, 2.76, 8.1 |
| CEO 23 | 6.37, 4.29, 7.56 | 6.06, 3.04, 8.72 | 5.78, 3.60, 7.06 | 6.80, 5.55, 7.95 | 5.01, 2.62, 6.88 |
| CEO 24 | 6.72, 4.95, 7.97 | 6.48, 3.75, 8.73 | 6.86, 4.39, 8.02 | 6.87, 5.92, 7.98 | 5.34, 2.68, 8.21 |
| CEO 25 | 6.86, 4.2, 8.05 | 5.64, 2.83, 7.67 | 6.67, 4.36, 7.82 | 6.77, 5.88, 8.04 | 5.45, 3.37, 6.93 |

**Figure A1**

*LLM instructions for demographic data only version.*

We are conducting a social experiment using a specific persona. Below are the detailed characteristics of this persona, including personality traits derived from the Moral Foundations Theory, emotional analysis, and demographic information. Please embody this persona and respond to all experimental questions accordingly. Do not incorporate your own opinions; instead, base your responses strictly on the provided data.

*Demographic information:*

- Age: [Insert Age]
- Gender: [Insert Gender]
- Education: [Insert Education]
- Professional Experience: [Insert Professional Experience]

Experiment Details

Experiment Name: Moral Foundations Questionnaire (MFQ30)

Objective: The primary objective of this study is to assess and measure individual differences in moral reasoning and moral values using the self-scorable Moral Foundations Questionnaire (MFQ30). The MFQ30 is designed to capture the extent to which various moral foundations are relevant to a person's judgments of right and wrong. It aims to provide insight into the relative importance of different moral considerations, such as harm, fairness, loyalty, authority, and purity, across diverse populations. The MFQ30 helps differentiate between individuals' moral priorities and correlates these differences with other psychological and behavioral measures. This tool is based on the Moral Foundations Theory, which proposes that people's moral reasoning is based on innate, modular foundations influenced by culture and individual experiences. The MFQ30 aims to provide a comprehensive understanding of how these foundations manifest in everyday moral judgments and decision-making.

Please read the following sentences and indicate your agreement or disagreement:

[0] = Strongly disagree

[1] = Moderately disagree

[2] = Slightly disagree

[3] = Slightly agree

[4] = Moderately agree

[5] = Strongly agree

Response Guidelines:

Respondents should consider the extent to which each moral foundation is relevant to their personal judgments about right and wrong.

Responses should reflect the persona's characteristics, values, and the extent to which these characteristics influence moral judgments.

Provide responses consistent with moral foundation theory

When responding, choose the most appropriate number on the 6-point scale to represent the persona's perspective on each statement.

Example Response Format:

Question: Whether or not someone suffered emotionally.

Response: (4) very relevant

Rationale: This response indicates that the persona places significant importance on the emotional impact of actions when making moral judgments.

**Figure A2**

*LLM instructions for textual data only version***.**

We are conducting a social experiment using a specific persona. Below are the characteristics of this persona from the file search. This also contains files in the file search option, they act as raw data to understand this person. It contains direct communications from this persona. Please embody this persona and respond to all experimental questions accordingly. Do not incorporate your own opinions; instead, base your responses strictly on the provided data.

Experiment Details

Experiment Name: Moral Foundations Questionnaire (MFQ30)

Objective: The primary objective of this study is to assess and measure individual differences in moral reasoning and moral values using the self-scorable Moral Foundations Questionnaire (MFQ30). The MFQ30 is designed to capture the extent to which various moral foundations are relevant to a person's judgments of right and wrong. It aims to provide insight into the relative importance of different moral considerations, such as harm, fairness, loyalty, authority, and purity, across diverse populations. The MFQ30 helps differentiate between individuals' moral priorities and correlates these differences with other psychological and behavioral measures. This tool is based on the Moral Foundations Theory, which proposes that people's moral reasoning is based on innate, modular foundations influenced by culture and individual experiences. The MFQ30 aims to provide a comprehensive understanding of how these foundations manifest in everyday moral judgments and decision-making.

Please read the following sentences and indicate your agreement or disagreement:

[0] = Strongly disagree

[1] = Moderately disagree

[2] = Slightly disagree

[3] = Slightly agree

[4] = Moderately agree

[5] = Strongly agree

Response Guidelines:

Respondents should consider the extent to which each moral foundation is relevant to their personal judgments about right and wrong.

Responses should reflect the persona's characteristics, values, and the extent to which these characteristics influence moral judgments.

Provide responses consistent with the raw data in the moral foundation theory and the file search to give the relevant option.

Give rational with both the MFT score and the file search option

When responding, choose the most appropriate number on the 6-point scale to represent the persona's perspective on each statement.

Example Response Format:

Question: Whether or not someone suffered emotionally.

Response: (4) Moderately agree

Rationale: This response indicates that the persona places significant importance on the emotional impact of actions when making moral judgments.

**Figure A3**

*LLM instructions for Moral Foundation Theory scores version*

We are conducting a social experiment using a specific persona. Below are the detailed characteristics of this persona, including personality traits derived from the Moral Foundations Theory, please embody this persona and respond to all experimental questions accordingly. Do not incorporate your own opinions; instead, base your responses strictly on the provided data.

**Persona Information:**

*Personality Traits Analysis*

*Moral Foundations Theory Values (Scale: 0–10):*

- Harm: Min: [Insert Min], Max: [Insert Max], Mean: [Insert Mean]
- Fairness: Min: [Insert Min], Max: [Insert Max], Mean: [Insert Mean]
- Loyalty: Min: [Insert Min], Max: [Insert Max], Mean: [Insert Mean]
- Authority: Min: [Insert Min], Max: [Insert Max], Mean: [Insert Mean]
- Purity: Min: [Insert Min], Max: [Insert Max], Mean: [Insert Mean]

Experiment Details

Experiment Name: Moral Foundations Questionnaire (MFQ30)

Objective: The primary objective of this study is to assess and measure individual differences in moral reasoning and moral values using the self-scorable Moral Foundations Questionnaire (MFQ30). The MFQ30 is designed to capture the extent to which various moral foundations are relevant to a person's judgments of right and wrong. It aims to provide insight into the relative importance of different moral considerations, such as harm, fairness, loyalty, authority, and purity, across diverse populations. The MFQ30 helps differentiate between individuals' moral priorities and correlates these differences with other psychological and behavioral measures. This tool is based on the Moral Foundations Theory, which proposes that people’s moral reasoning is based on innate, modular foundations influenced by culture and individual experiences. The MFQ30 aims to provide a comprehensive understanding of how these foundations manifest in everyday moral judgments and decision-making.

Please read the following sentences and indicate your agreement or disagreement:

[0] = Strongly disagree

[1] = Moderately disagree

[2] = Slightly disagree

[3] = Slightly agree

[4] = Moderately agree

[5] = Strongly agree

Response Guidelines:

Respondents should consider the extent to which each moral foundation is relevant to their personal judgments about right and wrong.

Responses should reflect the persona’s characteristics, values, and the extent to which these characteristics influence moral judgments.

Provide responses consistent with moral foundation theory

When responding, choose the most appropriate number on the 6-point scale to represent the persona's perspective on each statement.

Example Response Format:

Question: Whether or not someone suffered emotionally.

Response: (4) very relevant

Rationale: This response indicates that the persona places significant importance on the emotional impact of actions when making moral judgments.

**Figure A4**

*LLM instruction for Moral Foundation scores and textual data version.*

We are conducting a social experiment using a specific persona. Below are the characteristics of this persona from the moral foundation theory scores and the file search. This also contains files in the file search option; they act as raw data to understand this person. It contains direct communications from this persona.

Please embody this persona and respond to all experimental questions accordingly. Do not incorporate your own opinions; instead, base your responses strictly on the provided data.

**Persona Information:**

*Personality Traits Analysis*

*Moral Foundations Theory Values (Scale: 0–10):*

- Harm: Min: [Insert Min], Max: [Insert Max], Mean: [Insert Mean]
- Fairness: Min: [Insert Min], Max: [Insert Max], Mean: [Insert Mean]
- Loyalty: Min: [Insert Min], Max: [Insert Max], Mean: [Insert Mean]

- Authority: Min: [Insert Min], Max: [Insert Max], Mean: [Insert Mean]

- Purity: Min: [Insert Min], Max: [Insert Max], Mean: [Insert Mean]

Experiment Details

Experiment Name: Sacrificial Dilemma Moral Judgment Task

Objective:

The primary objective of this study is to evaluate individual differences in moral judgments related to sacrificial dilemmas. Specifically, the task explores participants' willingness to accept harm to a single individual to save multiple others. The dilemmas are designed to examine how personal and impersonal contexts influence moral acceptability judgments, providing insights into participants' ethical priorities, emotional engagement, and cognitive processing in morally conflicting situations. This experiment aims to correlate responses with broader measures of moral reasoning, empathy, and utilitarian decision-making.

Expected Output:

Responses on a 6-point scale with the following descriptors:

[1] = Absolutely unacceptable

[2] = Mostly unacceptable

[3] = Slightly unacceptable

[4] = Slightly acceptable

[5] = Mostly acceptable

[6] = Absolutely acceptable

Response Guidelines:

1. Respondents are asked to judge the moral acceptability of actions in sacrificial dilemmas.

2. Responses should reflect personal moral reasoning and prioritize either deontological ethics (focus on rules and individual rights) or utilitarian ethics (focus on outcomes and the greater good).

3. Provide a rationale for each response that includes the moral reasoning and considerations (e.g., harm, fairness, emotional proximity).

4. Align responses with the scenario's context (e.g., personal vs. impersonal involvement).

Example Response Format:

Question: Whether or not someone suffered emotionally.

Response: (4) Very relevant

Rationale: This response indicates that the persona places significant importance on the emotional impact of actions when making moral judgments.

**Figure A5**

*Moral Foundation Questionnaire*

***Part 1.*** *When you decide whether something is right or wrong, to what extent are the following considerations relevant to your thinking? Please rate each statement using this scale:*

[0] = not at all relevant (This consideration has nothing to do with my judgments of right and wrong)

[1] = not very relevant

[2] = slightly relevant

[3] = somewhat relevant

[4] = very relevant

[5] = extremely relevant (This is one of the most important factors when I judge right and wrong)

1. Whether or not someone suffered emotionally
2. Whether or not some people were treated differently than others
3. Whether or not someone's action showed love for his or her country
4. Whether or not someone showed a lack of respect for authority
5. Whether or not someone violated standards of purity and decency
6. Whether or not someone was good at math
7. Whether or not someone cared for someone weak or vulnerable
8. Whether or not someone acted unfairly
9. Whether or not someone did something to betray his or her group
10. Whether or not someone conformed to the traditions of society
11. Whether or not someone did something disgusting
12. Whether or not someone was cruel
13. Whether or not someone was denied his or her rights
14. Whether or not someone showed a lack of loyalty
15. Whether or not an action caused chaos or disorder
16. Whether or not someone acted in a way that God would approve of

***Part 2.*** *Please read the following sentences and indicate your agreement or disagreement:*

[0] = Strongly disagree

[1] = Moderately disagree

[2] = Slightly disagree

[3] = Slightly agree

[4] = Moderately agree

[5] = Strongly agree

17. Compassion for those who are suffering is the most crucial virtue.

18. When the government makes laws, the number one principle should be ensuring that everyone is treated fairly.

19. I am proud of my country's history.

20. Respect for authority is something all children need to learn.

21. People should not do things that are disgusting, even if no one is harmed.

22. It is better to do good than to do bad.

23. One of the worst things a person could do is hurt a defenseless animal.

24. Justice is the most important requirement for a society.

25. People should be loyal to their family members, even when they have done something wrong.

26. Men and women each have different roles to play in society.

27. I would call some acts wrong on the grounds that they are unnatural.

28. It can never be right to kill a human being.

29. I think it's morally wrong that rich children inherit a lot of money while poor children inherit nothing.

30. It is more important to be a team player than to express oneself.

31. If I were a soldier and disagreed with my commanding officer's orders, I would obey anyway because that is my duty.

32. Chastity is an important and valuable virtue.

To score the MFQ yourself, you can copy your answers into the grid below. Then add up the 6 numbers in each of the five columns and write each total in the box at the bottom of the column. The box then shows your score on each of 5 psychological "foundations" of morality. Scores run from 0-30 for each foundation. (Questions 6 and 22 are just used to catch people who are not paying attention. They don't count toward your scores.

**Figure A6**

*Scoreboard for Moral Foundations scores*

| Question # | Your Response | Question # | Your Response | Question # | Your Response | Question # | Your Response | Question # | Your Response | Question # | Your Response |
|---|---|---|---|---|---|---|---|---|---|---|---|
| **1** | | **2** | | **3** | | **4** | | **5** | | **6** | |
| **7** | | **8** | | **9** | | **10** | | **11** | | | |
| **12** | | **13** | | **14** | | **15** | | **16** | | | |
| **17** | | **18** | | **19** | | **20** | | **21** | | **22** | |
| **23** | | **24** | | **25** | | **26** | | **27** | | | |
| **28** | | **29** | | **30** | | **31** | | **32** | | | |

| Harm / Care | Fairness / Reciprocity | In-group/ Loyalty | Authority / Respect | Purity / Sanctity |
|---|---|---|---|---|
| | | | | |

**Figure A7**

*Sacrificial Dilemmas*

Likert Scale Description:

Responses on a 6-point scale with the following descriptors:

[1] = Absolutely unacceptable

[2] = Mostly unacceptable

[3] = Slightly unacceptable

[4] = Slightly acceptable

[5] = Mostly acceptable

[6] = Absolutely acceptable

The 6 questions to answer:

1. The Footbridge Problem: A runaway trolley is heading toward five people tied to the tracks. You are standing on a footbridge next to a very large individual. Pushing this individual off the bridge would stop the trolley and save the five people but would kill the individual. Would it be morally acceptable to push the individual?
2. The Hospital Organ Dilemma: Five patients in a hospital will die without organ transplants. A healthy individual with compatible organs walks into the hospital. Would it be morally acceptable to sacrifice this person to save the five patients?
3. The Switch Problem: A runaway trolley is heading toward five people tied to the tracks. You can pull a lever to divert the trolley onto another track where one person is tied. Would it be morally acceptable to pull the lever?
4. The Lifeboat Dilemma: You are on a lifeboat with six people, and the boat is sinking due to excess weight. To save the group, one person must be thrown overboard. Would it be morally acceptable to throw one person overboard?

5. The Burning Building Dilemma: You are a firefighter who has just entered a burning building. You can either save a single trapped child you see in front of you, or continue deeper into the building where you believe five adults are trapped but not visible. You do not have time to save both. Would it be morally acceptable to save the five adults instead of the one child?

6. The Refugee Dilemma: You are a border guard. Allowing a group of refugees to cross the border will save their lives, but doing so risks triggering violent retaliation from a hostile neighboring country, which could result in the deaths of several of your own citizens. Would it be morally acceptable to let the refugees cross?

**Figure A8**

*Sacrificial Dilemma instructions for LLMs.*

Goal: We are conducting a Social Experiment using a specific Persona. Below are the detailed characteristics of this persona, including personality traits derived from Demographic Information and The File Search tool.

IMPORTANT: Please embody these traits into this persona and respond to all experimental questions accordingly. Do not incorporate your own opinions; instead, base your responses strictly on the provided data.

Prompt Example: Expect these inputs in the user instruction along with using the file search option to access the raw data of this persona and adopt the persona.

*Demographic Information:*

Age: value

Gender: value

Education: value

Professional Experience: value

Experiment Details are as follows

Experiment Name: Sacrificial Dilemma Moral Judgment Task

Objective: The primary objective of this study is to evaluate individual differences in moral judgments related to sacrificial dilemmas. Specifically, the task explores participants' willingness to accept harm to a single individual to save multiple others. The dilemmas are designed to examine how personal and impersonal contexts influence moral acceptability judgments, providing insights into participants' ethical priorities, emotional engagement, and cognitive processing in morally conflicting situations. This experiment aims to correlate responses with broader measures of moral reasoning, empathy, and utilitarian decision-making.

Responses on a 6-point scale with the following descriptors:

[1] = Absolutely unacceptable

[2] = Mostly unacceptable

[3] = Slightly unacceptable

[4] = Slightly acceptable

[5] = Mostly acceptable

[6] = Absolutely acceptable

The 6 questions to answer:

1. The Footbridge Problem: A runaway trolley is heading toward five people tied to the tracks. You are standing on a footbridge next to a very large individual. Pushing this individual off the bridge would stop the trolley and save the five people but would kill the individual. Would it be morally acceptable to push the individual?

2. The Hospital Organ Dilemma: Five patients in a hospital will die without organ transplants. A healthy individual with compatible organs walks into the hospital. Would it be morally acceptable to sacrifice this person to save the five patients?

3. The Switch Problem: A runaway trolley is heading toward five people tied to the tracks. You can pull a lever to divert the trolley onto another track where one person is tied. Would it be morally acceptable to pull the lever?

4. The Lifeboat Dilemma: You are on a lifeboat with six people, and the boat is sinking due to excess weight. To save the group, one person must be thrown overboard. Would it be morally acceptable to throw one person overboard?

5. The Burning Building Dilemma: You are a firefighter who has just entered a burning building. You can either save a single trapped child you see in front of you, or continue deeper into the building where you believe five adults are trapped but not visible. You do not have time to save both. Would it be morally acceptable to save the five adults instead of the one child?

6. The Refugee Dilemma: You are a border guard. Allowing a group of refugees to cross the border will save their lives, but doing so risks triggering violent retaliation from a hostile neighboring country, which could result in the deaths of several of your own citizens. Would it be morally acceptable to let the refugees cross?

Response Guidelines:

1. Respondents are asked to judge the moral acceptability of actions in sacrificial dilemmas.

2. Responses should reflect personal moral reasoning and prioritize either deontological ethics (focus on rules and individual rights) or utilitarian ethics (focus on outcomes and the greater good).

4. Align responses with the scenario's context (e.g., personal vs. impersonal involvement).

Example Outputs formats:

1. Dilemma: The Footbridge Problem

Response: 2

2. Dilemma: The Switch Problem

Response: 5

3. The Hospital Organ Dilemma

Response: 3

4. The Lifeboat Dilemma

Response: 2

5. The Burning Building Dilemma

Response: 4

6. The Refugee Dilemma

Response: 3

SMAS Values = (2+5+3+2+4+3)/6 = 3.1666

The responses across all dilemmas will be averaged to create a Sacrificial Moral Acceptability

Score (SMAS). This score will help differentiate participants' reliance on utilitarian versus deontological principles in moral decision-making. Finally output the SMAS

**Table A3.1**

*Means, standard deviations, correlations, and regression coefficients for sacrificial judgments and moral foundations (Trait-Isolated Prompting, Run 1).*

| Item | M | SD | β (SE) | 1 | 2 | 3 | 4 | 5 |
|---|---|---|---|---|---|---|---|---|
| 1. Sacrifice Rating | 3.06 | 0.38 | – | – | | | | |
| 2. Harm | 4.78 | 0.5 | .17 (.08)* | .16* | – | | | |
| 3. Fairness | 4.66 | 0.54 | −.12 (.08) | −.09 | .17* | – | | |
| 4. Loyalty | 4.54 | 0.42 | −.03 (.08) | −.00 | .19* | .23** | – | |
| 5. Authority | 4.46 | 0.35 | .10 (.08) | 0.12 | 0.11 | −.07 | .16* | – |
| 6. Purity | 3.53 | 0.66 | .04 (.08) | 0.05 | 0.12 | 0.13 | 0.14 | 0.05 |

**Table A3.2**

*Means, standard deviations, correlations, and regression coefficients for sacrificial judgments and moral foundations (Trait-Isolated Prompting, Run 2).*

| Item | M | SD | β (SE) | 1 | 2 | 3 | 4 | 5 |
|---|---|---|---|---|---|---|---|---|
| 1. Sacrifice Rating | 2.98 | 0.31 | – | – | | | | |
| 2. Harm | 4.72 | 0.62 | −.06 (.08) | −.08 | – | | | |
| 3. Fairness | 4.63 | 0.48 | −.12 (.08) | −.10 | .21** | – | | |

| | | | | | | | | | |
|---|---|---|---|---|---|---|---|---|---|
| 4. Loyalty | 4.44 | 0.44 | .02 (.08) | 0.01 | 0.1 | .32*** | – | | |
| 5. Authority | 4.45 | 0.34 | .06 (.08) | 0.05 | 0.06 | .21** | .36*** | – | |
| 6. Purity | 3.55 | 0.80 | .06 (.08) | 0.04 | 0.09 | .18* | .19* | .16* | |

**Table A3.3**

*Means, standard deviations, correlations, and regression coefficients for sacrificial judgments and moral foundations (Trait-Isolated Prompting, Run 3).*

| Item | M | SD | β (SE) | 1 | 2 | 3 | 4 | 5 |
|---|---|---|---|---|---|---|---|---|
| 1. Sacrifice Rating | 2.98 | 0.32 | – | – | | | | |
| 2. Harm | 4.72 | 0.59 | −.01 (.08) | −.04 | – | | | |
| 3. Fairness | 4.50 | 0.44 | −.03 (.08) | −.05 | .20** | – | | |
| 4. Loyalty | 4.45 | 0.41 | .01 (.08) | −.02 | .30*** | .22** | – | |
| 5. Authority | 4.43 | 0.34 | .14 (.08) | 0.12 | 0.04 | 0.07 | .19* | – |
| 6. Purity | 3.50 | 0.69 | −.15 (.08) | −.14 | .18* | .22** | .27*** | 0.11 |

**Table A3.4**

*Means, standard deviations, correlations, and regression coefficients for sacrificial judgments and moral foundations (Trait-Isolated Prompting, Run 4).*

| Item | M | SD | β (SE) | 1 | 2 | 3 | 4 | 5 |
|---|---|---|---|---|---|---|---|---|
| 1. Sacrifice Rating | 2.98 | 0.32 | – | – | | | | |
| 2. Harm | 4.71 | 0.63 | −.04 (.08) | −.06 | – | | | |
| 3. Fairness | 4.47 | 0.48 | .04 (.08) | 0.02 | .17* | – | | |
| 4. Loyalty | 4.41 | 0.43 | −.04 (.08) | −.05 | .17* | .23** | – | |
| 5. Authority | 4.44 | 0.30 | .03 (.08) | 0.01 | 0.03 | .15* | 0.14 | – |
| 6. Purity | 3.42 | 0.70 | −.10 (.08) | −.11 | .20** | 0.12 | .16* | .18* |

**Table A3.5**

*Means, standard deviations, intercorrelations, and regression coefficients for sacrificial judgments and moral foundations (Trait-Isolated Prompting, Run 5).*

| Item | M | SD | β (SE) | 1 | 2 | 3 | 4 | 5 |
|---|---|---|---|---|---|---|---|---|
| 1. Sacrifice Rating | 2.98 | 0.32 | – | – | | | | |
| 2. Harm | 4.74 | 0.55 | −.04 (.08) | −.05 | – | | | |
| 3. Fairness | 4.49 | 0.46 | .07 (.08) | 0.03 | .18* | – | | |
| 4. Loyalty | 4.41 | 0.42 | .02 (.08) | 0.02 | 0.13 | .31*** | – | |
| 5. Authority | 4.45 | 0.36 | .07 (.08) | 0.07 | −.04 | .21** | .32*** | – |
| 6. Purity | 3.50 | 0.72 | −.21 (.08)** | −.19* | 0.13 | .26*** | .21** | 0.11 |

Note. N = 181 virtual CEOs. Sacrifice rating refers to the averaged acceptability judgments across six sacrificial dilemmas. Intercorrelations are presented below the diagonal. Regression coefficients (standard errors in parentheses) represent the unique predictive contribution of each moral foundation to sacrificial judgments, controlling for the others. $p < .05$, p < .01, $p < .001$.

**Table A4.1**

*OLS and Ridge Regression Coefficients for Trait-Isolated Prompting (Run-Level Results)*

| run | predictor | beta_ols | beta_ridge_lam1 |
|---|---|---|---|
| **1a** | Harm | 0.168 | 0.081 |
| **1a** | Fairness | -0.111 | -0.051 |
| **1a** | Loyalty | -0.028 | -0.008 |
| **1a** | Authority | 0.096 | 0.054 |
| **1a** | Purity | 0.043 | 0.023 |
| **1b** | Harm | -0.066 | -0.037 |
| **1b** | Fairness | -0.115 | -0.052 |
| **1b** | Loyalty | 0.021 | 0.008 |
| **1b** | Authority | 0.062 | 0.028 |
| **1b** | Purity | 0.053 | 0.023 |
| **1c** | Harm | -0.015 | -0.013 |
| **1c** | Fairness | -0.025 | -0.018 |
| **1c** | Loyalty | 0.004 | -0.003 |
| **1c** | Authority | 0.138 | 0.065 |
| **1c** | Purity | -0.148 | -0.07 |
| **1d** | Harm | -0.04 | -0.024 |
| **1d** | Fairness | 0.045 | 0.017 |
| **1d** | Loyalty | -0.041 | -0.021 |
| **1d** | Authority | 0.029 | 0.01 |
| **1d** | Purity | -0.106 | -0.053 |
| **1e** | Harm | -0.035 | -0.021 |
| **1e** | Fairness | 0.07 | 0.024 |
| **1e** | Loyalty | 0.026 | 0.012 |
| **1e** | Authority | 0.069 | 0.036 |
| **1e** | Purity | -0.217 | -0.1 |

Note. Regression coefficients (β) are reported for OLS and Ridge regression with λ = 1 across five separate runs (Tables 1a–1e). Predictors are the five moral foundations. Differences across runs illustrate variability in model estimates due to stochastic generation.

**Table A4.2**

*Coefficient Stability Summary for Trait-Isolated Prompting*

| | beta_ols_mean | beta_ols_sd | beta_ols_min | beta_ols_max | beta_ridge_mean | beta_ridge_sd |
|---|---|---|---|---|---|---|
| Authority | 0.079 | 0.041 | 0.029 | 0.138 | 0.039 | 0.021 |
| Fairness | -0.027 | 0.086 | -0.115 | 0.07 | -0.016 | 0.036 |
| Harm | 0.002 | 0.095 | -0.066 | 0.168 | -0.003 | 0.047 |
| Loyalty | -0.003 | 0.03 | -0.041 | 0.026 | -0.002 | 0.013 |
| Purity | -0.075 | 0.119 | -0.217 | 0.053 | -0.035 | 0.056 |

Note. Mean, standard deviation, minimum, and maximum values of regression coefficients across the five runs for each predictor. Also includes sign stability (count of runs with positive vs. negative coefficients).

**Table A4.3**

*Model Fit ($R^2$) for Trait-Isolated Prompting Across Runs*

| run | R2_ols | R2_ridge_lam1 |
|---|---|---|
| 1a | 0.051 | 0.025 |
| 1b | 0.022 | 0.011 |
| 1c | 0.039 | 0.019 |
| 1d | 0.017 | 0.009 |
| 1e | 0.05 | 0.024 |

Note. Model fit ($R^2$) reported separately for OLS and Ridge regression models across the five runs. Values reflect low but non-trivial explanatory power, comparable to findings in human samples.

**Table A5.1**

*Means, standard deviations, intercorrelations, and regression coefficients for sacrificial judgments and moral foundations (Integrated-Trait Prompting, Run 1).*

| Item | M | SD | β (SE) | 1 | 2 | 3 | 4 | 5 |
|---|---|---|---|---|---|---|---|---|
| 1. Sacrifice Rating | 2.98 | 0.31 | – | – | | | | |
| 2. Harm | 4.72 | 0.6 | −.11 (.08) | −.10 | – | | | |
| 3. Fairness | 4.4 | 0.48 | −.02 (.08) | −.02 | .20** | – | | |
| 4. Loyalty | 4.46 | 0.42 | .04 (.08) | 0.02 | .17* | .30*** | – | |
| 5. Authority | 4.52 | 0.41 | −.04 (.08) | −.04 | 0.06 | .19* | .28*** | – |
| 6. Purity | 3.53 | 0.73 | .04 (.08) | 0.03 | .18* | .20** | .27*** | 0.1 |

**Table A5.2**

*Means, standard deviations, intercorrelations, and regression coefficients for sacrificial judgments and moral foundations (Integrated-Trait Prompting, Run 2).*

| Item | M | SD | β (SE) | 1 | 2 | 3 | 4 | 5 |
|---|---|---|---|---|---|---|---|---|
| 1. Sacrifice Rating | 3 | 0.31 | – | – | | | | |
| 2. Harm | 4.6 | 0.91 | −.10 (.09) | −.05 | – | | | |
| 3. Fairness | 4.27 | 1.24 | .06 (.09) | 0.04 | .35*** | – | | |
| 4. Loyalty | 4.48 | 0.62 | .04 (.10) | 0.04 | .45*** | .32*** | – | |
| 5. Authority | 4.46 | 0.65 | .07 (.10) | 0.06 | .38*** | .30*** | .63*** | – |
| 6. Purity | 3.32 | 1.04 | −.02 (.09) | 0 | .36*** | .46*** | .38*** | .29*** |

**Table A5.3**

*Means, standard deviations, intercorrelations, and regression coefficients for sacrificial judgments and moral foundations (Integrated-Trait Prompting, Run 3).*

| Item | M | SD | β (SE) | 1 | 2 | 3 | 4 | 5 |
|---|---|---|---|---|---|---|---|---|
| 1. Sacrifice Rating | 3 | 0.31 | – | – | | | | |
| 2. Harm | 4.53 | 0.92 | .03 (.09) | 0.04 | – | | | |
| 3. Fairness | 4.27 | 1.24 | .02 (.09) | 0.03 | .37*** | – | | |
| 4. Loyalty | 4.46 | 0.62 | .02 (.10) | 0.02 | .47*** | .30*** | – | |
| 5. Authority | 4.4 | 0.65 | −.04 (.09) | −.01 | .38*** | .15* | .58*** | – |
| 6. Purity | 3.31 | 1.07 | .00 (.09) | 0.02 | .37*** | .45*** | .27*** | .16* |

**Table A5.4**

*Means, standard deviations, intercorrelations, and regression coefficients for sacrificial judgments and moral foundations (Integrated-Trait Prompting, Run 4).*

| **Item** | **M** | **SD** | **β (SE)** | **1** | **2** | **3** | **4** | **5** |
|---|---|---|---|---|---|---|---|---|
| 1. Sacrifice Rating | 3 | 0.31 | – | – | | | | |
| 2. Harm | 4.58 | 0.88 | .01 (.08) | 0.02 | – | | | |
| 3. Fairness | 4.3 | 1.21 | .05 (.09) | 0.03 | .33*** | – | | |
| 4. Loyalty | 4.49 | 0.52 | .08 (.09) | 0.05 | .34*** | .22** | – | |
| 5. Authority | 4.47 | 0.51 | −.06 (.09) | −.02 | .28*** | .24** | .53*** | – |
| 6. Purity | 3.35 | 1.07 | −.04 (.08) | −.02 | .27*** | .40*** | .15* | .22** |

**Table A5.5**

*Means, standard deviations, intercorrelations, and regression coefficients for sacrificial judgments and moral foundations (Integrated-Trait Prompting, Run 5).*

| Item | M | SD | β (SE) | 1 | 2 | 3 | 4 | 5 |
|---|---|---|---|---|---|---|---|---|
| 1. Sacrifice Rating | 3 | 0.31 | – | – | | | | |
| 2. Harm | 4.58 | 0.91 | .02 (.09) | 0.03 | – | | | |
| 3. Fairness | 4.24 | 1.23 | .04 (.09) | 0.04 | .38*** | – | | |
| 4. Loyalty | 4.4 | 0.6 | .11 (.12) | 0.03 | .50*** | .32*** | – | |
| 5. Authority | 4.44 | 0.57 | −.13 (.11) | −.04 | .41*** | .32*** | .72*** | – |
| 6. Purity | 3.21 | 1.04 | −.02 (.09) | 0.01 | .35*** | .48*** | .34*** | .31*** |

Note. N = 181 virtual CEOs. Sacrifice rating refers to the averaged acceptability judgments across six sacrificial dilemmas. Intercorrelations are presented below the diagonal. Regression coefficients (standard errors in parentheses) represent the unique predictive contribution of each moral foundation to sacrificial judgments, controlling for the others. $p < .05$, p < .01, $p < .001$.

**Table A6.1:**

*OLS and Ridge Regression Coefficients for Integrated-Trait Prompting (Run-Level Results)*

| run | predictor | beta_ols | beta_ridge_lam1 |
|---|---|---|---|
| IIa | Harm | -0.11 | -0.052 |
| IIa | Fairness | -0.011 | -0.007 |
| IIa | Loyalty | 0.043 | 0.016 |
| IIa | Authority | -0.048 | -0.021 |
| IIa | Purity | 0.045 | 0.019 |
| IIb | Harm | -0.103 | -0.037 |
| IIb | Fairness | 0.054 | 0.021 |
| IIb | Loyalty | 0.034 | 0.017 |
| IIb | Authority | 0.067 | 0.029 |
| IIb | Purity | -0.021 | -0.006 |
| IIc | Harm | 0.04 | 0.018 |
| IIc | Fairness | 0.016 | 0.011 |
| IIc | Loyalty | 0.019 | 0.007 |
| IIc | Authority | -0.039 | -0.012 |
| IIc | Purity | -0.001 | 0.004 |
| IId | Harm | 0.008 | 0.007 |
| IId | Fairness | 0.04 | 0.016 |
| IId | Loyalty | 0.079 | 0.028 |
| IId | Authority | -0.066 | -0.019 |
| IId | Purity | -0.036 | -0.014 |
| IIe | Harm | 0.018 | 0.013 |
| IIe | Fairness | 0.049 | 0.02 |
| IIe | Loyalty | 0.109 | 0.021 |
| IIe | Authority | -0.137 | -0.033 |
| IIe | Purity | -0.014 | 0.0 |

Note. Regression coefficients (β) are reported for OLS and Ridge regression with λ = 1 across five separate runs (Tables 1a–1e). Predictors are the five moral foundations. Differences across runs illustrate variability in model estimates due to stochastic generation.

**Table A6.2**

*Coefficient Stability Summary for Integrated-Trait Prompting*

| predictor | beta_ols_mean | beta_ols_sd | beta_ols_min | beta_ols_max | beta_ridge_mean | beta_ridge_sd |
|---|---|---|---|---|---|---|
| Authority | -0.0446 | 0.07330279 | -0.137 | 0.067 | -0.0112 | 0.02371 |
| Fairness | 0.0296 | 0.02698703 | -0.011 | 0.054 | 0.0122000 | 0.01143 |
| Harm | -0.0294 | 0.07137086 | -0.11 | 0.04 | -0.01020 | 0.03199 |
| Loyalty | 0.0568 | 0.03659508 | 0.019 | 0.109 | 0.0178 | 0.007661 |
| Purity | -0.0054 | 0.03087555 | -0.036 | 0.045 | 0.000600 | 0.01232 |

Note. Mean, standard deviation, minimum, and maximum values of regression coefficients across the five runs for each predictor. Also includes sign stability (count of runs with positive vs. negative coefficients).

**Table A6.3**

*Model Fit ($R^2$) for Integrated-Trait Prompting Across Runs*

| run | R2_ols | R2_ridge_lam1 |
|---|---|---|
| IIa | 0.015 | 0.007 |
| IIb | 0.013 | 0.005 |
| IIc | 0.003 | 0.001 |
| IId | 0.007 | 0.003 |
| IIe | 0.011 | 0.003 |